\documentclass[review]{elsarticle}

\usepackage{amsmath,amsfonts,amssymb}
\usepackage{subcaption}
\usepackage{float}
\usepackage{microtype}
\usepackage{booktabs}
\usepackage{mathtools}

\usepackage{lineno,hyperref}

\journal{Journal of Artificial Intelligence In Medicine}

\begin{document}
\nolinenumbers

\begin{frontmatter}

\title{The Concordance Index Decomposition: A Measure for a Deeper Understanding of Survival Prediction Models
}



\author[ca]{Abdallah Alabdallah}\corref{mycorrespondingauthor}
\cortext[mycorrespondingauthor]{Corresponding author}
\ead{abdallah.alabdallah@hh.se}

\author[ca,lu]{Mattias Ohlsson}
\author[ca,ri]{Sepideh Pashami}
\author[ca]{Thorsteinn Rögnvaldsson}


\address[ca]{Center for Applied Intelligent Systems Research (CAISR), Halmstad University, Sweden}
\address[lu]{Dept. Astronomy and Theoretical Physics, Lund University, Sweden}
\address[ri]{RISE Research Institutes of Sweden, Stockholm, Sweden}

\begin{abstract}
The Concordance Index (C-index) is a commonly used metric in Survival Analysis for evaluating the performance of a prediction model. In this paper, we propose a decomposition of the C-index into a weighted harmonic mean of two quantities: one for ranking observed events versus other observed events, and the other for ranking observed events versus censored cases. This decomposition enables a finer-grained analysis of the relative strengths and weaknesses between different survival prediction methods.
The usefulness of this decomposition is demonstrated through benchmark comparisons against classical models and state-of-the-art methods, together with the new variational generative neural-network-based method (SurVED) proposed in this paper. The performance of the models is assessed using four publicly available datasets with varying levels of censoring. 
Using the C-index decomposition and synthetic censoring, the analysis shows that deep learning models utilize the observed events more effectively than other models. This allows them to keep a stable C-index in different censoring levels. In contrast to such deep learning methods, classical machine learning models deteriorate when the censoring level decreases due to their inability to improve on ranking the events versus other events.

\end{abstract}


\end{frontmatter}


\section{Introduction}
\label{sec:introduction}
More and more data is being collected to improve the estimation of the probability of survival and the expected remaining lifetime, for humans as well as equipment. Making such estimates is the purpose of Survival Analysis. This is an analysis of the time to an \emph{event}, e.g., an individual’s death or the breakdown of a piece of equipment. While several statistical methods for survival analysis have been developed~\cite{Kleinbaum:2012}, the availability of large quantities of data has spurred the development of machine learning (ML) based approaches that consider more intricate covariate effects~\cite{Wang:2019}.

An important aspect of survival analysis is handling \emph{censored} cases, e.g., hospitalized patients who do not experience a relapse before the end of a study, equipment that is replaced before a breakdown, or equipment that has not experienced a breakdown yet. Censoring is very common in clinical studies and can occur for various reasons. It is possible for a patient not to experience the event during the study’s timeframe (for example, death or relapse). Also, a patient might experience a different event, making it impossible to follow up on the event of interest. 

Censoring also makes it more difficult to evaluate the goodness-of-fit when the target variable is not fully observed. Several evaluation metrics have been proposed to assess various aspects of a model's performance \cite{Rahman:2017}. However, the Concordance Index (C-index) is one of the most used metrics as it encompasses both observed events and censored cases. In doing so, it quantifies the rank correlation between actual survival times and a model's predictions. Multiple C-index estimators have been proposed, like Harrel's C-index~\cite{Harrell:1982}, Uno's C-index~\cite{Uno:2011} (a modified weighted version of Harrel's C-index), and Gonen and Heller's measure~\cite{Gonen:2005}. The latter serves as an alternative estimator based on the reversed definition of concordance. Finally, a time-dependent version of the C-index was proposed in~\cite{Antolini:2005}, which takes the whole survival function into consideration.

Harrel's C-index, the focus of this study, is perhaps the most often used index and has an intuitive and straightforward interpretation. It measures the ability of a predictor to order subjects by estimating the proportion of correctly ordered pairs among all comparable pairs in the dataset. In the presence of censoring, there are two types of times; event time and censoring time. This results in two types of comparable pairs: event vs. event $(ee)$ and event vs. censored $(ec)$. A predictor may not perform equally well in ranking both types of comparable pairs. Comparisons of given models' performance using the C-index tend to show few significant differences in those datasets with a high ratio of censored cases. More significant differences however appear on datasets with low censoring ratios. This phenomenon can be attributed to unseen differences in the models' abilities to rank the different types of pairs $(ee)$ and $(ec)$.

We therefore propose a decomposition of the C-index into a weighted harmonic mean of two quantities: the C-index for ranking observed events ($CI_{ee}$), and a C-index for ranking observed events versus censored cases ($CI_{ec}$), weighted by $\alpha \in [0, 1]$. This decomposition makes it easier to understand an algorithm’s strengths and weaknesses under different censoring levels. As such, the role of the weighting factor $\alpha$ in assessing the balance of a predictor when dealing with the two categories of pairs, namely $(ee)$ and $(ec)$ becomes clearer.

From a modeling perspective, the primary outcome of such survival analyses is the Survival Function denoted as $S(t)=P(T > t)$, which represents the probability of surviving beyond time $t$, where $T$ is the event time. Over time a number of classical statistical and machine learning models have been developed to estimate the survival function $S(t)$ in a non-parametric, semi-parametric, or parametric way~\cite{Kaplan:1958,Wei:1992,Cox:1972,Ishwaran:2008,VanBelle:2010,VanBelle:2011}. More recently however, deep learning models have been introduced for survival time modeling~\cite{Ranganath:2016,Katzman:2018,Lee_2018,Chapfuwa:2018,Miscouridou:2018,JING:2019,Xiu:2020,Nagpal:2021a,Hu:2021,XU:2023}. DeepSurv~\cite{Katzman:2018}, for example, is a direct extension of the Cox Proportional Hazard (CPH) model~\cite{Cox:1972} that employs a deep neural network in place of the CPH linear predictor. As such, DeepSurv maintains the constraint of the proportional hazards assumption. Unlike DeepSurv however, some deep learning models discretize the survival timeline. Most notably, DeepHit~\cite{Lee_2018} estimates the probability mass function based on a discrete output. Predictions from such discrete-time models, in contrast to continuous-time models, are however constrained by the choice of the upper limit of the output timeline.

Deep generative models facilitate the estimation of date distributions. In the case of survival analysis, deep generative models can be utilized to estimate the distribution of the event times in both parametric and non-parametric ways ~\cite{Ranganath:2016,Miscouridou:2018}. The Deep Adversarial Time-to-Event model (DATE)~\cite{Chapfuwa:2018} for example, is a survival model based on Generative Adversarial Networks (GAN)~\cite{Goodfellow:2014}. DATE estimates the event distribution in a non-parametric manner using adversarial training and is trained to generate $p(t|\mathbf{x})$ while penalizing fake samples ($\mathbf{x}, t$). However, such GAN models suffer from instability issues, such as the Mode Collapse and the Non-Convergence problems, making them challenging to train and potentially lead to a poor local equilibrium~\cite{kodali:2017, Chen:2021}.

Recently, the Variational Survival Inference (VSI) model~\cite{Xiu:2020} was introduced, adopting variational inference to approximate $p(t|\mathbf{x})$. VSI is a discrete-time model that employs two encoders, $p(z|\mathbf{x})$ and $q(z|\mathbf{x},t)$, and encourages these two distributions to be similar by using Kullback-Leibler divergence which means the model can better account for interactions between covariates and event times. In addition, the VSI model discretized output constrains the prediction timeline to be limited by the maximum time in the training data. To highlight the importance of the interactions between the covariates and the event times captured by the $q$ branch, the authors of the VSI model developed a variant of VSI, labeled VSI-NoQ which lacks the encoder’s q branch. It is worth noting that although the VSI performs significantly better than VSI-NoQ, the role of the $q(z|\mathbf{x},t)$ branch is unclear. 

In this work, a new survival model is proposed: SurVED (Survival Variational Encoder-Decoder). SurVED is essentially a translation of the Variational Auto Encoder (VAE)~\cite{Kingma:2014} into the field of survival analysis. It is a conditional generative model with a single encoder and a single decoder, which learns to model the distribution of events conditioned on the covariates $\mathbf{x}$. 

SurVED and VSI are both variational-inference-based models. However, SurVED derives its objective function from the DATE model~\cite{Chapfuwa:2018}. This adaptation enables SurVED to deal with continuous time where, unlike the VSI model, no discretization is required. Moreover, SurVED does not impose any upper-limit constraint on the timeline of the model predictions. The loss function has separate terms with different weights for censored and non-censored samples. Additionally, SurVED and VSI differ in terms of architecture. Specifically, while VSI comprises two encoders $p(\mathbf{z}|\mathbf{x})$ and $q(\mathbf{z}|\mathbf{x}, t)$, where $q$ is utilized to capture the interactions between the covariates and the event times, SurVED uses only one encoder. This makes SurVED more similar to the variant VSI-NoQ, albeit with additional regularization on the latent space, continuous output, and a different loss function.

In summary, this work presents two contributions. Firstly, it derives a decomposition of the concordance index which provides insights into the distinctions between seemingly similar-performing models. It also helps to understand why there are larger-magnitude differences between classical and deep learning models in the case of low censoring. Ultimately, by showing areas of strengths and weaknesses, the C-index decomposition has the potential to serve as a guide in the development of new survival models and offers insights to enhance existing ones. Additionally, this work introduces a new continuous-time variational-based model that overcomes the limitations of its predecessors, DATE and VSI, and achieves a ranking performance comparable to the state of the art.  


\section{Method}
\label{sct:method}
In this section, we introduce the Concordance Index Decomposition as a new approach to highlight the difference between survival models. Additionally, we present the SurVED model (Survival Variational Encoder-Decoder) and provide an overview of the four datasets used for numerical tests and comparisons.

\subsection{The Concordance Index Decomposition}
\label{ssct:The Concordance Index Decomposition}
The C-index is a measure of the probability that the predicted event times ($\hat{t}_i$, $\hat{t}_j$) of two randomly selected subjects maintain the same relative order as their true event times ($t_i$, $t_j$), i.e., $P(\hat{t}_i > \hat{t}_j | t_i > t_j)$. It's important to note that not all pairs can be compared when censoring is present; a pair $(\mathbf{x}_i,\mathbf{x}_j)$ is comparable (usable) if the earliest time represents an event, or both times are events. Conversely, a pair is deemed not comparable if the earliest time is censored or if both are censored cases~\cite{Harrell:1996}.

The C-index can be decomposed into two parts; one to measure the relative ordering of cases with observed events, and another to measure the ordering of cases with observed events relative to censored cases. This decomposition is useful when comparing how methods perform in situations with a high proportion of censored cases, to situations with a low proportion of censored cases.

We define the random variable $o_{ij}=\hat{t}_i > \hat{t}_j | t_i > t_j$ that takes the value $1$ if the $ij$ pair is ordered (concordant) and $0$ if it is discordant. We also define the random variable $k_{ij}$, which takes the value ($1$) if the ($ij$) pair is an event-event $(ee)$ pair and ($0$) if the ($ij$) is an event-censored $(ec)$ pair. To simplify the notation, $P(o)$ represents $P(o_{ij}=1)$, $P(ee)$ represents $P(k_{ij}=1)$, and $P(ec)$ represents $P(k_{ij}=0)$. Note that $P(ee) +P(ec)=1$. With these definitions, the C-index can be written as $CI = P(o)$, and hence:
\begin{eqnarray}
    \frac{1}{CI} &=& \frac{1}{P(o)}
\nonumber \\
    &=& \frac{P(ee)+P(ec)}{P(o)}
\nonumber \\
    &=& \frac{P(ee)}{P(o)} + \frac{P(ec)}{P(o)}
\nonumber \\
    &=& \frac{P(o|ee)}{P(o|ee)}  \frac{P(ee)}{P(o)} + \frac{P(o|ec)}{P(o|ec)}  \frac{P(ec)}{P(o)}
\nonumber \\
    &=& \frac{P(o|ee)P(ee)}{P(o)} \frac{1}{P(o|ee)} + \frac{P(o|ec)P(ec)}{P(o)} \frac{1}{P(o|ec)}
\nonumber \\
    &=& P(ee|o) \frac{1}{P(o|ee)} + P(ec|o) \frac{1}{P(o|ec)}
\nonumber \\
    &=& P(ee|o) \frac{1}{P(o|ee)} + (1-P(ee|o)) \frac{1}{P(o|ec)}
\nonumber
\end{eqnarray}
We define $CI_{ee}$ as a C-index for event-event cases, $CI_{ec}$ as a C-index for events-censored cases, and we introduce the notation $\alpha$ for the conditional probability that the pair is an event-event pair ($ee$) given that it is a correctly ordered pair:
\begin{eqnarray}
    CI_{ee} & \equiv & P(o|ee)
\label{eq:P_CI_ee} \\
    CI_{ec} & \equiv & P(o|ec)
\label{eq:P_CI_ec} \\
    \alpha & \equiv & P(ee|o) = 1 - P(ec|o)
\label{eq:P_alpha}
\end{eqnarray}
This yields the following relationship, which shows that the full C-index ($CI$) is a weighted harmonic mean of the C-indices defined for the subsets $ee$ and $ec$: 
\begin{equation}
    \frac{1}{CI} = \alpha \frac{1}{CI_{ee}} + (1 - \alpha) \frac{1}{CI_{ec}}
\label{eq:CI_ee_ec}
\end{equation}

The C-index and its decomposed parts $CI_{ee}$, $CI_{ec}$, and $\alpha$ can be estimated based on the number of correctly ordered pairs $N^+$, incorrectly ordered pairs $N^-$, and the number of ties $N^=$. Since there are two kinds of comparable (usable) pairs: event-event pairs ($ee$) and event-censored pairs ($ec$), then:
\begin{eqnarray}
    N^+ = N_{ee}^+ + N_{ec}^+
\nonumber \\
    N^- = N_{ee}^- + N_{ec}^-
\nonumber \\
    N^= = N_{ee}^= + N_{ec}^=
\label{eq:N}
\end{eqnarray}
There are multiple ways to handle ties, and we use the Somers' $d$ measure~\cite{Somers:1962}, which considers the ties in the event cases to be incomparable pairs. It also considers the ties in the predicted values to be binary random guesses; hence, half of them are counted as correctly ordered. 
\begin{equation}
    CI =  \frac{N^+ + \frac{1}{2} N^=}{N^+ + N^- + N^=} 
	= \frac{N_{ee}^+ + N_{ec}^+ + \frac{1}{2} N_{ee}^= + \frac{1}{2}  N_{ec}^=}{N_{ee}^+ + N_{ec}^+ + N_{ee}^- + N_{ec}^- + N_{ee}^= + N_{ec}^=}
\label{eq:CI_N}
\end{equation}
From expressions~(\ref{eq:P_CI_ee}), (\ref{eq:P_CI_ec}), and (\ref{eq:P_alpha}) we thus have:
\begin{eqnarray}
    CI_{ee} & = & \frac{N_{ee}^+ + \frac{1}{2} N_{ee}^=}{N_{ee}^+ + N_{ee}^- + N_{ee}^=}
\label{eq:CIee_N} \\
    CI_{ec} & = & \frac{N_{ec}^+ + \frac{1}{2} N_{ec}^=}{N_{ec}^+ + N_{ec}^- + N_{ec}^=}
\label{eq:CIec_N} \\
    \alpha & = & \frac{N_{ee}^+ + \frac{1}{2} N_{ee}^=}{N_{ee}^+ + N_{ec}^+ + \frac{1}{2} N_{ee}^= + \frac{1}{2} N_{ec}^=}
\label{eq:alpha_N}
\end{eqnarray}

The factor $\alpha$ is the conditional probability that the pair is event-event ($ee$) given that it is a correctly ordered pair. This factor weights the contribution of the correct ordering of event-event pairs relative to the correct ordering of event-censored pairs in the C-index. Changes in $\alpha$ are directly associated with variations in the model's performance in accurately ordering pairs and indirectly related to the ratio of observed events to censored cases in the dataset. A predictor that can order all events and censored cases correctly will have an $\alpha$ value equal to the fraction of event-event pairs within the comparable pairs, a value we can denote as $\alpha^{*}$. However, even an imperfect predictor can have $\alpha = \alpha^{*}$ as long as it scores equally on event-event pairs and event-censored pairs in proportion to their percentages; such a predictor can be denoted as a ``balanced'' predictor. 

The $\alpha$-Deviation is defined as the difference between $\alpha$ and $\alpha^{*}$. A predictor that excels at ordering event-event ($ee$) pairs more than event-censored ($ec$) pairs will have $\alpha > \alpha^{*}$, resulting in a positive $\alpha$-Deviation. On the other hand, a predictor that is better at ordering event-censored ($ec$) pairs compared to event-event ($ee$) pairs will have $\alpha < \alpha^{*}$, leading to a negative $\alpha$-Deviation.

\begin{eqnarray}
    \mbox{$\alpha$-Deviation} & \equiv & \alpha - \alpha ^{*}
 \label{alpha-dev} \\
    \alpha^{*} & \equiv & \frac{N_{ee}}{N_{ee} + N_{ec}}
 \label{alpha-opt}
\end{eqnarray}
where $N_{ee}$ and $N_{ec}$ are the number of the comparable (ee) and (ec) pairs in the dataset.
In this paper, we study the absolute value of the $\alpha$-Deviation. This is a measure of how unbalanced the predictor is when making mistakes.

\subsection{SurVED: Survival Variational Encoder-Decoder}
\label{ssct:SurVED: Survival Variational Encoder-Decoder}
Our model, SurVED, employs a conditional generator $G_\theta$ to estimate $f(t|\mathbf{x})$, the distribution of death conditioned on the covariate vector $\mathbf{x}$, with $\theta$ representing the parameters of the model. This generative model can be sampled to produce the conditional death function $f(t|\mathbf{x})$, from which the conditional cumulative death distribution function ($F$) and the conditional survival function ($S$) can be computed:
\begin{eqnarray}
	F(t|\mathbf{x}) & = & \int_{0}^{t} f(\tau|\mathbf{x}) \,d\tau
\label{eq:F} \\
    S(t|\mathbf{x}) & = & 1 - F(t|\mathbf{x})
\label{eq:S}
\end{eqnarray}

The model comprises two components: an Encoder $E_{\theta_1}(\mathbf{z}|\mathbf{x})$ which encodes the input $\mathbf{x}$ into a multi-dimensional Gaussian latent space represented by $(\mu_z, \sigma_z)$, and a Decoder $D_{\theta_2}(t|\mathbf{z})$ responsible for decoding a sample $\mathbf{z}$ from the latent space and generating a sample $t$ from the conditional distribution $f(t|\mathbf{x})$. Here $\theta_1$ and $\theta_2$ constitute $\theta$; the total parameters of $G_{\theta}$. For each input $\mathbf{x}$, $n$ values $t_i$ ($i = 1, \ldots, n$) from $f(t|\mathbf{x})$ are sampled. The survival function can be estimated using the Kaplan-Meier estimator considering the sampled times $t_i$ as observed event times. These $n$ samples ($t_i$) are also utilized to estimate the expected value $\mathbb{E}_{t \sim f(t|\mathbf{x})}[t]$ for the purpose of model evaluation.

\subsubsection{The Objective Function}
\label{sssct:The Objective Function}
The objective function of the generative model $G_{\theta}$ consists of four parts: $L_e$, $L_c$, $L_{KL}$, and $C_{lb}$. The first two, $L_e$ and $L_c$, represent construction losses that are separately evaluated for event cases and censored cases. These losses are designed to optimize the balance between events and censored cases. The third term, $L_{KL}$, originates from the VAE formulation and is the Kullback-Leibler divergence, serving as a regularization term.
The first three terms are: 
\begin{eqnarray}
    L_e & = & \mathbb{E}_{\mathbf{x} \sim 
    P_e(\mathbf{x})} \left[ | t - G_{\theta}(\mathbf{x}) | \right]
    \label{eq:L_e} \\
    L_c & = & \mathbb{E}_{\mathbf{x} \sim P_c(\mathbf{x})} \left[
    \max(0,t-G_{\theta}(\mathbf{x})) \right]
    \label{eq:L_c} \\
    L_{KL} & = &
    \mbox{\it KL} \left( P(\mathbf{z}|\mathbf{x}), N(0,1) \right)
    \label{eq:L_kl}
\end{eqnarray}
where the subscripts, \emph{e} and \emph{c}, indicate that the terms exclusively involve event cases or censored cases, respectively. The notation $P_e(\mathbf{x})$ denotes that $\mathbf{x}$ was drawn from the event cases, while $P_c(\mathbf{x})$ indicates that $\mathbf{x}$ was drawn from the censored cases. Additionally, $\mbox{\it KL}(p,q)$ represents the Kullback-Leibler divergence between the two distributions $p$ and $q$.
The fourth term 
\begin{equation}
	C_{lb}(\theta,\varepsilon) = 
	\frac{1}{|\varepsilon|}  \sum_{(\mathbf{x}_i, \mathbf{x}_j) \in \varepsilon} \left( 1 + \frac{\log \sigma(G_{\theta}(\mathbf{x}_i) - G_{\theta}(\mathbf{x}_j))}{\log 2} \right) 
\label{eq:cindexlb}
\end{equation}
is a differentiable lower bound for the C-index~\cite{Steck:2008}. Here, $\varepsilon$ is the set of comparable pairs, the symbol $\sigma$ is the standard sigmoid function, and $| \varepsilon |$ denotes the set $\varepsilon$ cardinality. 
Adding the $C_{lb}$ term to the loss function enables the model to directly optimize the C-index, encouraging concordance in the model predictions. The SurVED model aims to minimize the total loss: 
\begin{equation}
    L = \lambda_e L_e + \lambda_c L_c + \lambda_{KL} L_{KL} - \lambda_{lb} C_{lb}
\label{eq:total_loss}
\end{equation}
where the $\lambda_e$, $\lambda_c$, $\lambda_{KL}$, and $\lambda_{lb}$ are tunable weights. 

These objective terms have been used previously in the literature in different settings. The $L_e$ and $L_c$ terms, eqs.~(\ref{eq:L_e}) and (\ref{eq:L_c}), match the $\ell_2$ and $\ell_3$ terms used in the DATE loss function ~\cite{Chapfuwa:2018}. However, they can be traced back to earlier work by Van~Belle et al.~\cite{VanBelle:2010}. The fourth objective term, eq.~(\ref{eq:cindexlb}), was suggested for the DATE model~\cite{Chapfuwa:2018} as well.

\subsection{Description of Datasets}
\label{ssct:datasets_description}
The SurVED method has been evaluated against the reference methods on four publicly available medical datasets. The datasets are all fairly large, and cover different censoring levels, number of samples, and number of features; see Table~\ref{tbl:data_sets_description}. They have also been used in several previous benchmark studies.

FLCHAIN: A dataset used in a study~\cite{Dispenzieri:2012} to determine whether the free light chain (FLC) assay is a predictor of better/worse survival for the general population. The study showed that a high FLC was significantly predictive of worse overall survival. 

METABRIC: The Molecular Taxonomy of Breast Cancer International Consortium dataset~\cite{Bilal:2013}. This dataset is used to predict the survivability of breast cancer patients using gene expression profiles and clinical data.

NWTCO: Data from the US National Wilm's Tumor Study to predict survival based on tumor histology~\cite{Breslow:1999}. This data is available in the package \texttt{survival} in R~\cite{survival-package}.

SUPPORT: This data comes from the Study to Understand Prognoses and Preferences for Outcomes and Risks of Treatment~\cite{Knaus:1995}. This study aimed to understand the survival of seriously ill hospitalized patients and validate the predictions of a new prognostic model against an existing prognostic model and predictions by physicians. The SUPPORT data is sometimes split into subsets since there is more than one diagnosis, but it is used as one dataset here.

\begin{table}[H]
\caption{Dataset statistics}
\label{tbl:data_sets_description}
\vskip 0.15in
\begin{center}
\begin{small}
\begin{sc}
\begin{tabular}{lcccc}
\toprule
Dataset & Events ($\%$) & Samples & Features & Missing Values ($\%$) \\
\midrule
FLCHAIN  & 27.55$\%$ & 7,874 & 25 & 0.6$\%$ \\
METABRIC & 44.83$\%$ & 1,981 & 79 & 0.0$\%$ \\
NWTCO    & 14.18$\%$ & 4,028 & 9 & 0.0$\%$ \\
SUPPORT  & 68.11$\%$ & 9,105 & 59 & 6.5$\%$ \\
\bottomrule
\end{tabular}
\end{sc}
\end{small}
\end{center}
\vskip -0.1in
\end{table}

\subsection{Experimental Settings}

Seven models were compared: Cox Proportional Hazard model (CPH), Random Survival Forest (RSF), Deep Adversarial Time-to-Event model (DATE), DeepHit, DeepSurv, Variational Learning of Individual Survival Distributions model (VSI), and our model Survival Variational Encoder-Decoder (SurVED). The models were first compared based on the C-index performance and then analyzed further using the C-index Decomposition. 

The same sampling scheme was applied to all the experiments: $30\%$ of the data was used as a hold-out test set, and the remaining $70\%$ was used for hyperparameter tuning and training.  
The models were tuned using five-fold cross-validation, maximizing the C-index performance. At each fold, three sets were used for training, one set for early stopping for deep learning models, and the last set was used for validation. The early stopping set was not used for optimizing RSF.
In the final testing phase, a $100$-fold testing on the hold-out test set was done, varying the training data. At each fold, $90\%$ of the training data was used to train the models keeping $10\%$ as a validation set for deep-learning models.

The categorical features were one-hot encoded, and the numerical features were standardized with zero mean and unit variance. The target variable was scaled by the maximum value of the training set, and power transformed. Moreover, the missing values were filled with the training data median and mode for numerical and categorical features, respectively. The deep learning models were configured with a common architecture that included two hidden layers with $32$ nodes, a hyperbolic tangent activation function, and a $0.5$ dropout rate on the first hidden layer.
For the models that have special types of structure (DATE and VSI), we used the suggested structures in their repositories. SurVED has a latent size of four nodes and a single-layer linear perceptron as its decoder. Details about the network structures, data standardization, and transformation are available on our Github repository\footnote{https://github.com/abdoush/SurVED}. DATE's implementation from the authors' GitHub repository\footnote{https://github.com/paidamoyo/adversarial\_time\_to\_event} was used, while the Scikit-Survival library~\cite{sksurv} was used for the CPH and the RSF models. Moreover, the VSI model implementation provided by the authors on Github\footnote{https://github.com/ZidiXiu/VSI} was used. For DeepHit and DeepSurv, the PyCox library~\cite{Kvamme:2019} was used. A random search was performed to optimize the weights of the loss functions for DeepHit and SurVED. The number of output bins for the two discrete models, VSI and DeepHit, were optimized with choices including $[100, 200, 400, 1000]$. Additionally, a random search was conducted for RSF to optimize parameters such as \texttt{max\_depth}, \texttt{min\_samples\_split}, \texttt{min\_samples\_leaf}, and \texttt{max\_features}.

\section{Results and Discussion}
\label{sct:results}

\subsection{Comparison on the four data sets}
\label{ssct:Models Comparison}

Tables~\ref{tbl:comparison_with_other_models}, \ref{tbl:comparison_with_other_models_CID_Cee}, \ref{tbl:comparison_with_other_models_CID_Cec}, and \ref{tbl:comparison_with_other_models_CID_a_div}, present a comprehensive list of methods' scores based on the C-index ($CI$), C-index for event-event pairs ($CI_{ee}$), C-index for event-censored pairs ($CI_{ec}$), and $\alpha$-Deviation on the four datasets. In the context of the C-index, higher values indicate better performance. Conversely, when considering the $\alpha$-Deviation, lower values reflect a more ``balanced'' model, i.e., it performs more equally in ordering event-event and event-censored pairs. The statistical significance level was set to $5\%$, and hypothesis testing was carried out with $100$-fold test values using the Wilcoxon rank-sum test.

\begin{figure}[ht!]
     \centering
     \begin{subfigure}[b]{0.5\textwidth}
         \centering
         \includegraphics[width=\textwidth]{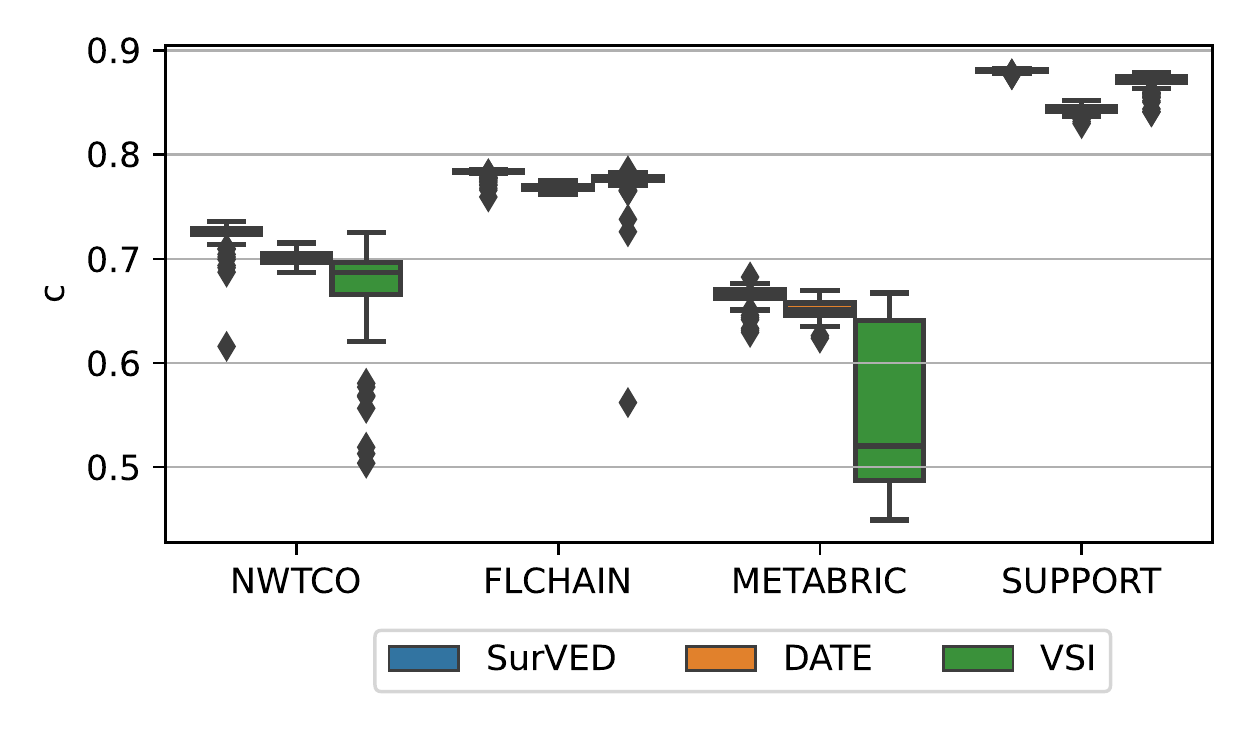}
         \vspace{-1.8\baselineskip}
         \caption{C-index}
         \label{fig:C}
     \end{subfigure}%
     ~
     \begin{subfigure}[b]{0.5\textwidth}
         \centering
         \includegraphics[width=\textwidth]{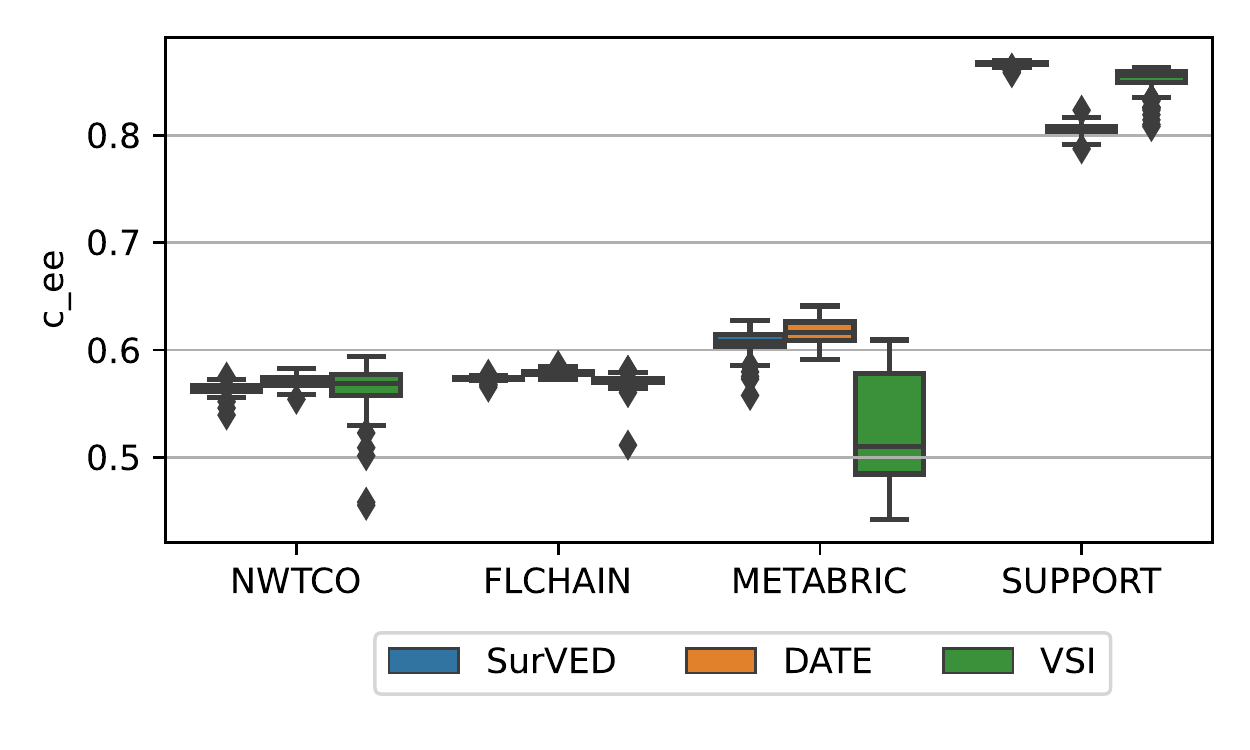}
         \vspace{-1.8\baselineskip}
         \caption{C$_{ee}$}
         \label{fig:C_ee}
     \end{subfigure}%
     \hfill
     \begin{subfigure}[b]{0.5\textwidth}
         \centering
         \includegraphics[width=\textwidth]{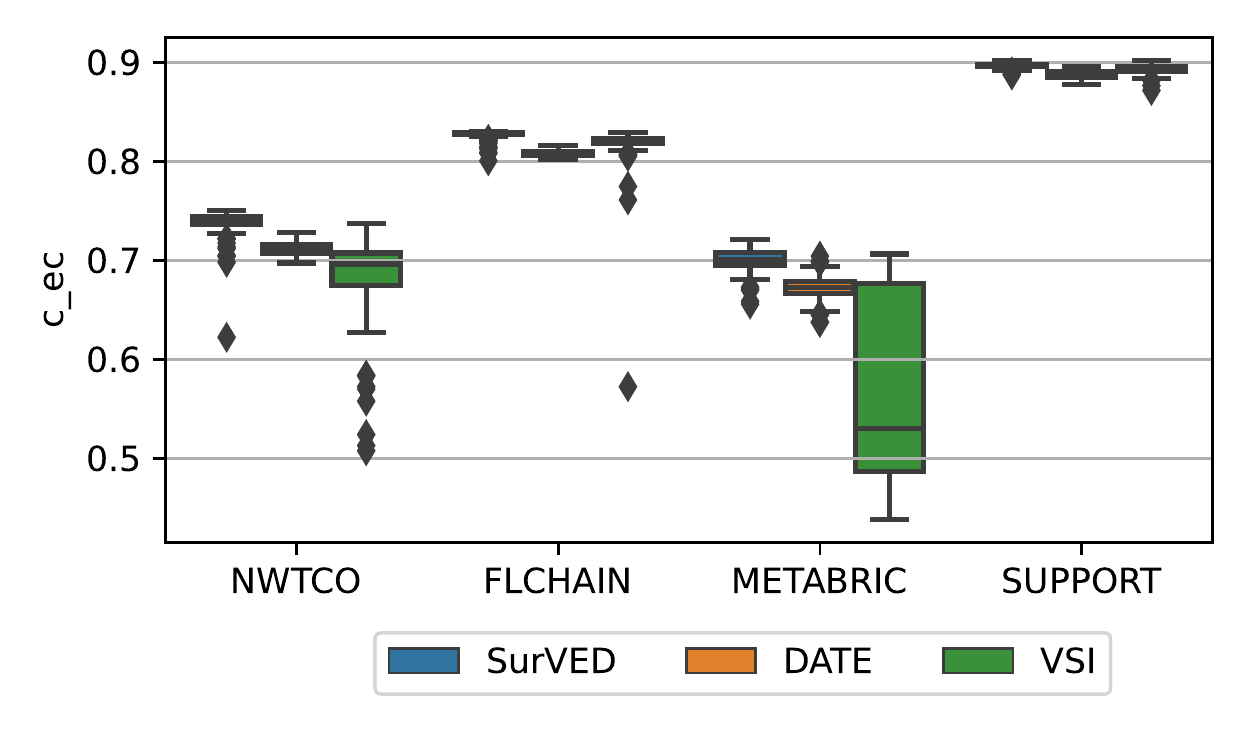}
         \vspace{-1.8\baselineskip}
         \caption{C$_{ec}$}
         \label{fig:C_ec}
     \end{subfigure}%
     ~
     \begin{subfigure}[b]{0.5\textwidth}
         \centering
         \includegraphics[width=\textwidth]{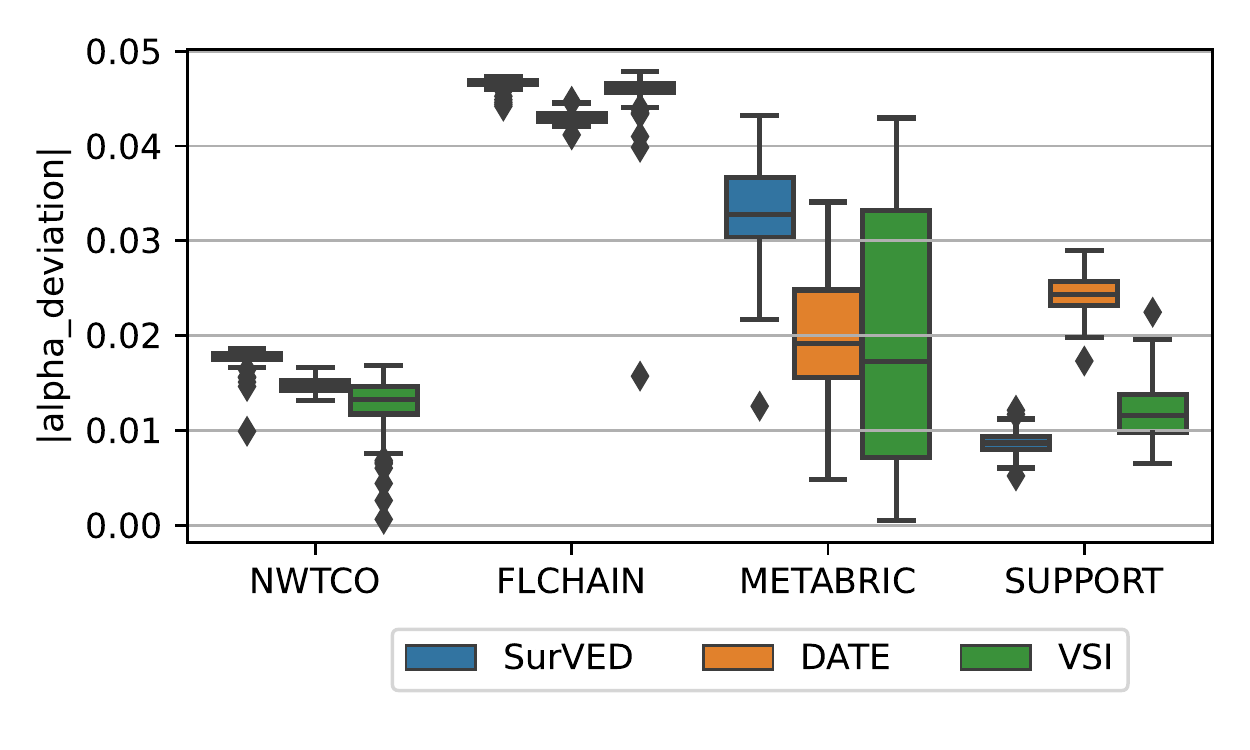}
         \vspace{-1.8\baselineskip}
         \caption{$\alpha$-Deviation}
         \label{fig:Alpha-Dev}
     \end{subfigure} 
        \caption{The results of $CI$, $\alpha$, $CI_{ee}$, $CI_{ec}$, and $|\alpha\text{-Deviation}|$ in eq.~(\ref{eq:CI_ee_ec}) of the SurVED, DATE, and VSI models on the four datasets (Censoring level decreases from the highest censoring (NWTCO) to the lowest censoring (SUPPORT)).}
        \label{fig:CID}
\end{figure}

We begin by comparing SurVED with DATE and VSI as it has close ties to both models. SurVED shares the same loss function with the DATE model and employs a variational inference approach similar to the VSI model. The results depicted in Figure~\ref{fig:CID} demonstrate that SurVED, with its regression-based loss function, outperformed the discrete-time-based VSI and the GAN-based DATE model across all datasets. 

It is worth noting that the VSI model exhibited unstable performance on METABRIC datasets, depicted by the large variance of its results as shown in Figure~\ref{fig:CID}. This instability may be attributed to the fact that METABRIC is the smallest dataset with the largest time horizons spanning over $9,200$ days. In such cases, time discretization can lead to information loss.

Remarkably, although SurVED outperformed DATE in the C-index on NWTOC, FLCHAIN, and METABRIC they demonstrated contrasting behaviors regarding $CI_{ee}$, $CI_{ec}$, and $\alpha$-Deviation. While DATE showed better performance in $CI_{ee}$ on these three datasets, SurVED was better in terms of $CI_{ec}$. Additionally, due to its higher $\alpha$-Deviation, SurVED placed higher weight on the $CI_{ec}$, resulting in a higher overall $CI$ performance. 

Looking at the full list of results in Tables~\ref{tbl:comparison_with_other_models}, \ref{tbl:comparison_with_other_models_CID_Cee}, \ref{tbl:comparison_with_other_models_CID_Cec}, and \ref{tbl:comparison_with_other_models_CID_a_div} we see that in the cases where there are no significant differences between the models in the C-index, they show significant differences in the decomposition terms $CI_{ee}$ and $CI_{ec}$.

For example, comparing RSF and DeepHit on the NWTCO dataset shows that RSF has a significantly better $CI_{ee}$ with no significant difference observed on the $CI_{ec}$. However, because DeepHit has a higher $\alpha$-Deviation, it places more weight on the $CI_{ec}$, resulting in no significant difference in the overall C-index. A similar scenario unfolds when comparing SurVED and CPH on the FLCHAIN dataset.

More interesting cases show contrasting differences in the decomposition terms leading to an insignificant difference in the C-index due to weighted averaging. For instance, on the NWTCO dataset, DeepHit exhibits a higher $CI_{ee}$ while CPH outperforms in $CI_{ec}$. Consequently, the total C-index shows no significant difference. A similar phenomenon is observed on the FLCHAIN dataset when comparing RSF with DeepHit and DeepSurv, where RSF excels in $CI_{ee}$ while DeepHit and DeepSurv demonstrate better performance in $CI_{ec}$, thereby diminishing the difference in the total C-index. This pattern is also observed in the comparison between DeepHit and DeepSurv on the FLCHAIN and the METABRIC datasets.

Contrasting differences in the decomposition terms do not always diminish the difference in the total C-index. In some cases, a higher $\alpha$-Deviation can outweigh one model over another. For example, consider the comparison of SurVED and DeepSurv on NWTCO, where DeepSurv exhibits a higher $CI_{ee}$ while SurVED has a higher $CI_{ec}$. Nevertheless, SurVED's higher $\alpha$-Deviation shifts the balance in favor of the $CI_{ec}$ term, resulting in a higher C-index. Similar scenarios arise in various cases like the comparison of CPH with RSF, DATE, VSI, and DeepSurv on NWTCO. In all these cases CPH demonstrates a lower $CI_{ee}$ but a higher $CI_{ec}$ and a higher $\alpha$-Deviation resulting in a higher C-index.

Occasionally, outweighing one term does not compensate for the differences in the terms, especially when the difference is substantial. For example, consider the case of CPH compared to DATE, DeepHit, and DeepSurv on the METABRIC dataset. While CPH has a higher $CI_{ec}$ and a higher $\alpha$-Deviation, it has a much lower $CI_{ee}$. In this scenario, outweighing the $CI_{ec}$ term does not compensate for the considerable gap in the $CI_{ee}$ term, resulting in CPH having a significantly lower total C-index. 

Poor performance on the METABRIC dataset was observed for the DeepHit model. This is similar to the VSI model which shares the discrete-time property with DeepHit. It is worth noting that this result cannot be compared to the result reported in DeepHit paper~\cite{Lee_2018} as they used a different version of the METABRIC dataset, where they re-scaled the time step to a month instead of a day as in our case. Additionally, they used the time-dependent C-index (C$_{td}$) as an evaluation measure.

Overall, the results indicate that classical models either outperformed or performed equally well compared to deep learning models for the smaller datasets with higher censoring levels. RSF was the best on METABRIC, while CPH was the best on NWTCO. On FLCHAIN, RSF shares the best performance with DeepSurv and DeepHit. However, deep learning models have a clear advantage on SUPPORT, the largest dataset with the lowest censoring level.

\begin{figure}[ht!]
     \centering
     \begin{subfigure}[b]{0.5\textwidth}
         \centering
         \includegraphics[width=\textwidth]{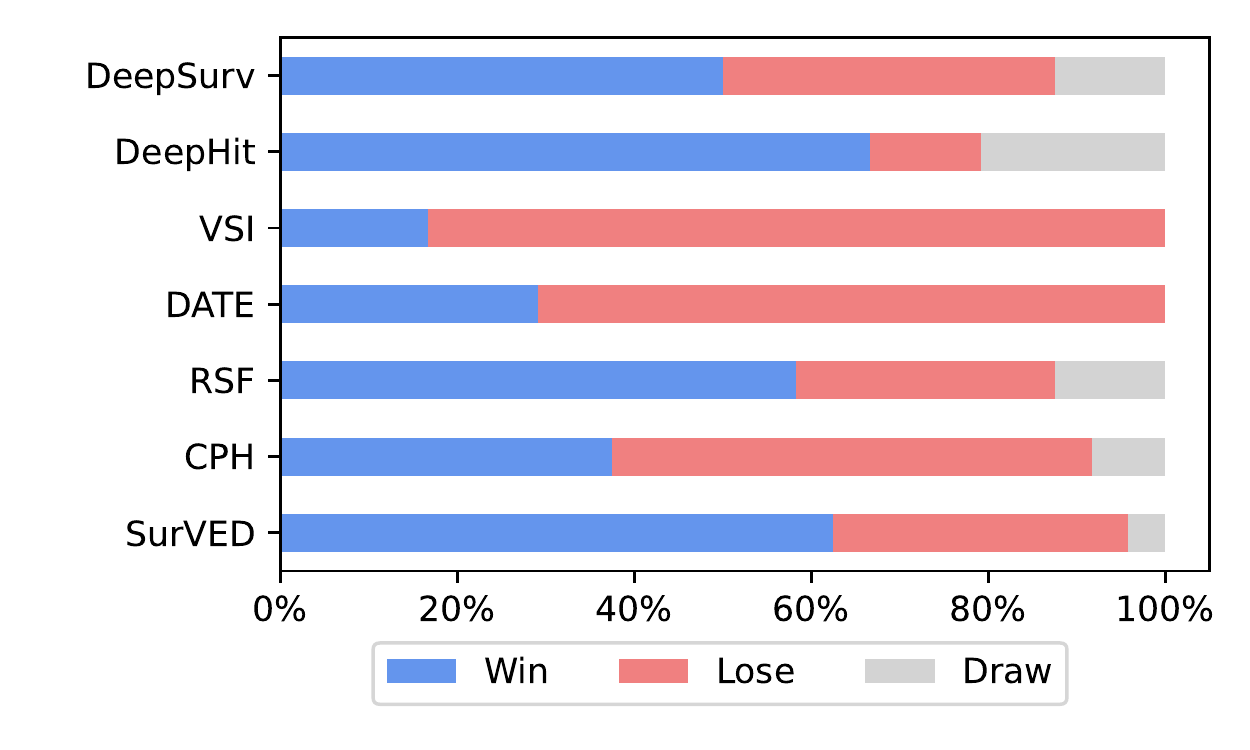}
         \caption{CI}
         \label{fig:wld_C}
     \end{subfigure}%
     ~
     \begin{subfigure}[b]{0.5\textwidth}
         \centering
         \includegraphics[width=\textwidth]{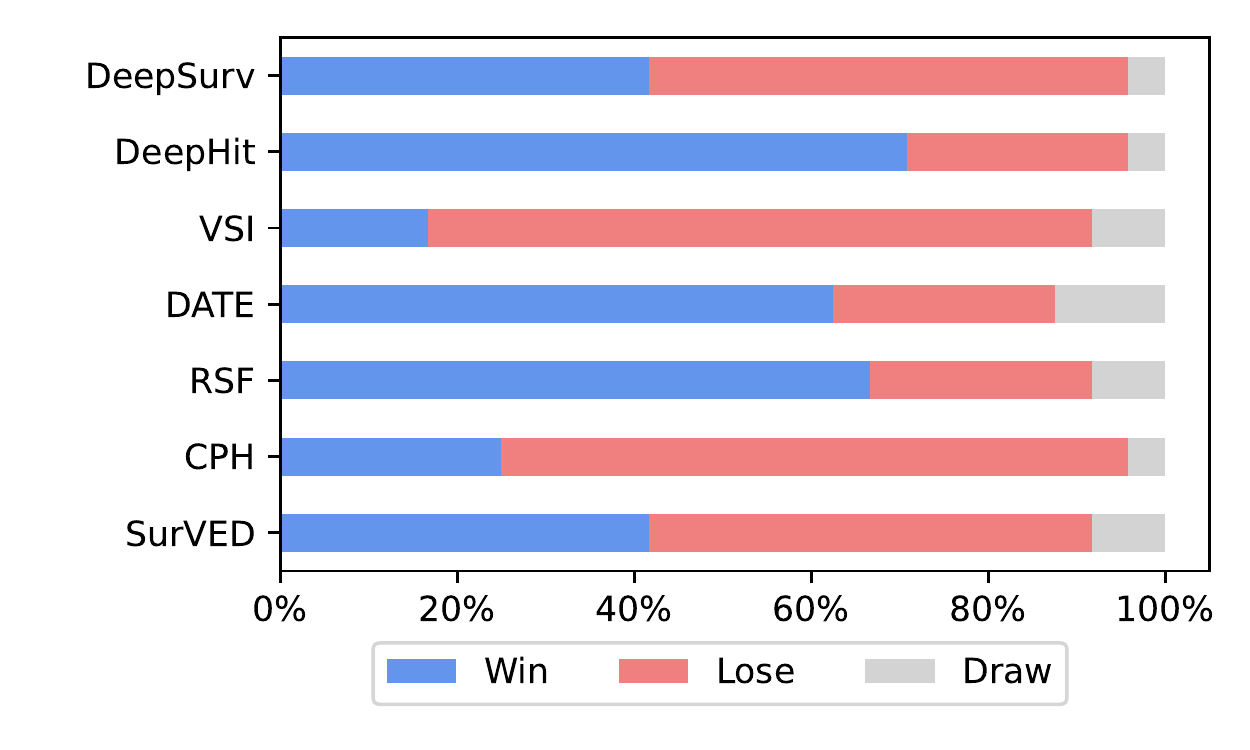}
         \caption{CI$_{ee}$}
         \label{fig:wld_C_ee}
     \end{subfigure}%
     
     \begin{subfigure}[b]{0.5\textwidth}
         \centering
         \includegraphics[width=\textwidth]{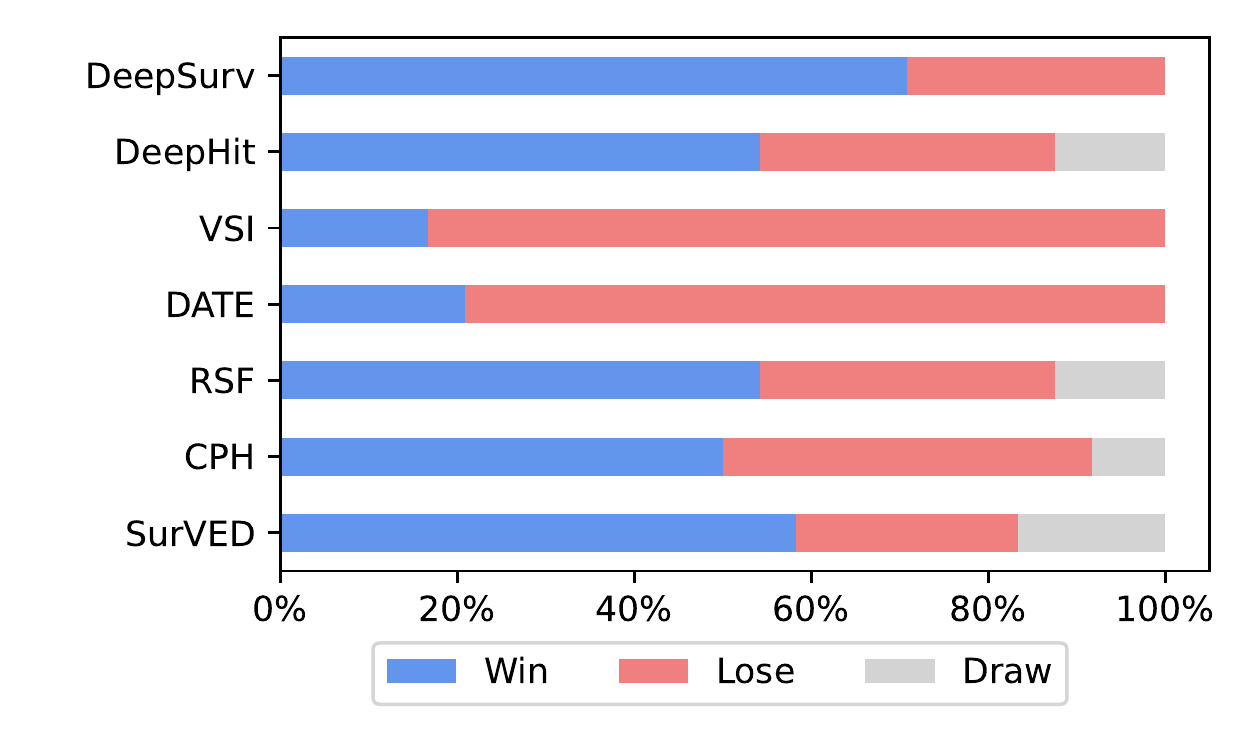}
         \caption{CI$_{ec}$}
         \label{fig:wld_C_ec}
     \end{subfigure}%
     ~
     \begin{subfigure}[b]{0.5\textwidth}
         \centering
         \includegraphics[width=\textwidth]{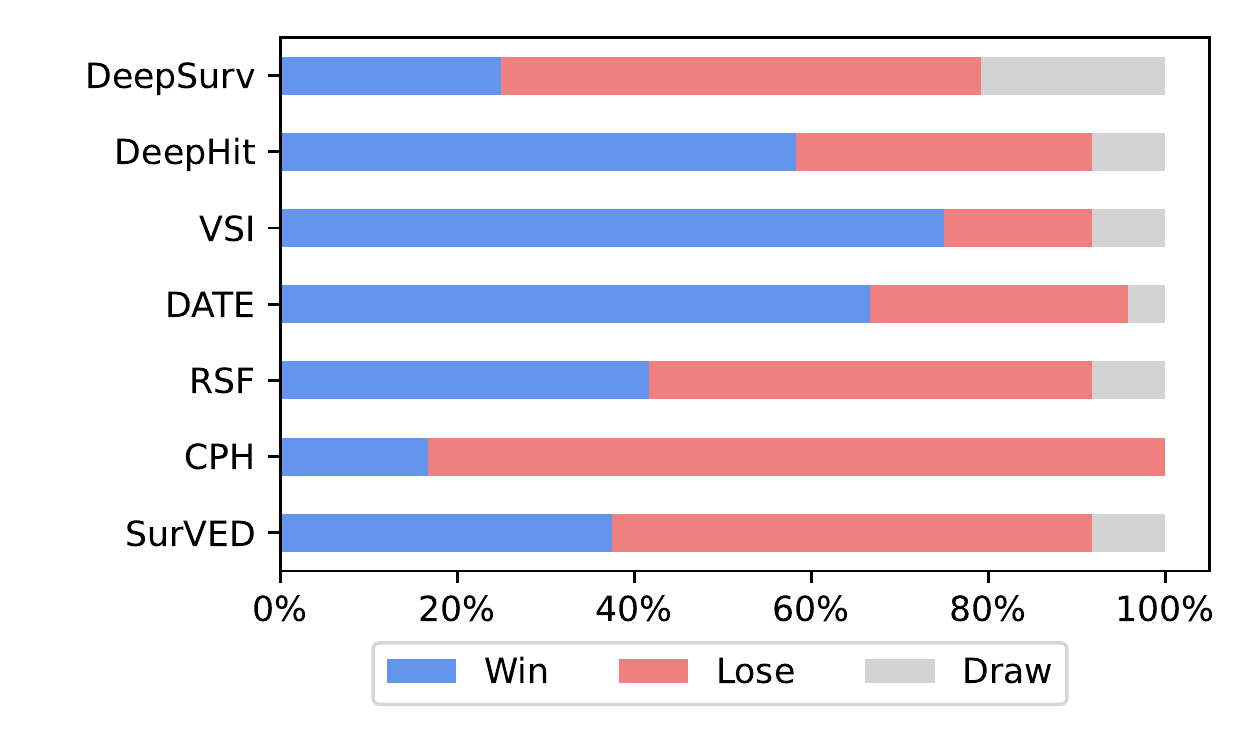}
         \caption{$\alpha$-Deviation}
         \label{fig:wld_Alpha-Dev}
     \end{subfigure}
        \caption{The Win/Lose/Draw comparison based on $CI$, $CI_{ee}$, $CI_{ec}$, and $\alpha$-Deviation in eq.~(\ref{eq:CI_ee_ec}) of the compared models on the four datasets.}
        \label{fig:WinLoseDraw}
\end{figure}

To assess the models comprehensively, pair-wise comparisons were performed between the seven models on the four datasets. Each model was compared against the other six models on each dataset, resulting in $24$ comparisons for each model. The results are summarized in Figure~\ref{fig:WinLoseDraw} as Win/Lose/Draw.

The chi-square test was applied to the Win/Lose/Draw data in Figure~\ref{fig:WinLoseDraw}, treating draws as a $50\%$ chance of winning or losing. 
Regarding the C-index performances, SurVED, DeepHit, RSF, and DeepSurv show a similar performance, whereas CPH, DATE, and VSI lag behind.
However, analyzing the other C-index decomposition terms reveals more interesting insights. For example, DATE has an excellent performance in terms of the $CI_{ee}$ but falls short in the $CI_{ec}$ which impacts its overall C-index. In contrast, the VSI model shows poor performance in both terms. The results also show that the main differences between the models stem from the $CI_{ee}$ part, while all models, except for DATE and VSI, exhibit similar overall $CI_{ec}$ performance. 

\begin{table}[t!]
\caption{The C-index ($CI$) values (\%) of the compared models on the four datasets. Numbers show the median, the 2.5\%, and the 97.5\% quantiles of 100-folds. The highest numerical value in each dataset is boldfaced.
}
\label{tbl:comparison_with_other_models}
\vskip 0.15in
\begin{center}
\begin{footnotesize}
\begin{sc}
\resizebox{\textwidth}{!}{%
\begin{tabular}{lcccc}
\toprule
& NWTCO & FLCHAIN & METABRIC & SUPPORT  \\
\midrule
CPH      & \textbf{72.91 (72.57, 73.25)} & 78.37 (78.30, 78.46) & 63.90 (63.02, 64.68) & 84.29 (84.19, 84.57) \\
RSF      & 72.84 (72.23, 73.40) & 78.43 (78.28, 78.62) & \textbf{67.80 (67.22, 68.49)} & 84.17 (83.80, 84.55)\\ 
DATE     & 70.06 (68.85, 71.32) & 76.84 (76.44, 77.38) & 65.09 (63.49, 66.87) & 84.38 (83.54, 84.96) \\
DeepSurv & 72.05 (70.40, 73.24) & \textbf{78.45 (77.99, 78.58)} & 64.40 (61.96, 66.11) & 87.88 (87.55, 88.05) \\
DeepHit  & 72.88 (70.43, 73.36) & 78.43 (78.27, 78.57) & 63.99 (63.26, 64.80) & \textbf{88.22 (88.01, 88.42)} \\
VSI      & 68.69 (53.68, 71.35) & 77.70 (75.09, 78.24) & 52.04 (45.61, 65.94) & 87.40 (84.71, 87.77) \\
SurVED   & 72.75 (69.26, 73.37) & 78.40 (76.92, 78.53) & 66.63 (63.72, 67.58) & 88.13 (87.76, 88.27) \\ 
\bottomrule
\end{tabular}}
\end{sc}
\end{footnotesize}
\end{center}
\vskip -0.1in
\end{table}
\begin{table}[t!]
\caption{The $CI_{ee}$ values (\%) of the compared models on the four datasets. Numbers show the median, the 2.5\%, and the 97.5\% quantiles of 100-folds. The highest numerical value in each dataset is boldfaced.}
\label{tbl:comparison_with_other_models_CID_Cee}
\vskip 0.15in
\begin{center}
\begin{footnotesize}
\begin{sc}
\resizebox{\textwidth}{!}{%
\begin{tabular}{lcccc}
\toprule
& NWTCO & FLCHAIN & METABRIC & SUPPORT  \\
\midrule
CPH      & 56.38 (56.08, 56.65) & 57.48 (57.37, 57.58) & 55.98 (54.84, 56.99) & 82.28 (82.14, 82.78) \\
RSF      & \textbf{57.39 (56.60, 57.94)} & \textbf{57.90 (57.73, 58.08)} & 61.43 (60.62, 62.29) & 80.54 (79.83, 81.28) \\ 
DATE     & 57.16 (56.04, 57.94) & 57.89 (57.45, 58.45) & \textbf{61.63 (59.34, 63.86)} & 80.58 (79.24, 81.58) \\
DeepSurv & 56.67 (55.46, 57.64) & 57.43 (56.96, 57.60) & 57.78 (56.66, 59.12) & 85.94 (85.56, 86.19) \\
DeepHit  & 57.28 (55.95, 57.72) & 57.59 (57.47, 57.71) & 59.17 (57.87, 60.40) & \textbf{86.82 (86.63, 86.98)} \\
VSI      & 56.90 (50.49, 58.52) & 57.15 (56.40, 57.97) & 51.03 (46.01, 60.15) & 85.63 (81.67, 86.21) \\
SurVED   & 56.36 (55.36, 57.18) & 57.37 (56.84, 57.62) & 60.83 (57.74, 62.03) & 86.70 (86.08, 86.94) \\ 
\bottomrule
\end{tabular}}
\end{sc}
\end{footnotesize}
\end{center}
\vskip -0.1in
\end{table}
\begin{table}[t!]
\caption{The $CI_{ec}$ values (\%) of the compared models on the four datasets. Numbers show the median, the 2.5\%, and the 97.5\% quantiles of 100-folds. The highest numerical value in each dataset is boldfaced.}
\label{tbl:comparison_with_other_models_CID_Cec}
\vskip 0.15in
\begin{center}
\begin{footnotesize}
\begin{sc}
\resizebox{\textwidth}{!}{%
\begin{tabular}{lcccc}
\toprule
& NWTCO & FLCHAIN & METABRIC & SUPPORT  \\
\midrule
CPH      & \textbf{74.34 (73.97, 74.71)} & 82.79 (82.70, 82.89) & 68.70 (67.73, 69.75) & 86.56 (86.36, 86.71) \\
RSF      & 74.18 (73.56, 74.76) & 82.76 (82.60, 82.99) & \textbf{71.75 (71.03, 72.58)} & 88.30 (88.16, 88.47) \\ 
DATE     & 71.19 (69.88, 72.57) & 80.86 (80.40, 81.53) & 67.31 (64.54, 69.63) & 88.79 (87.94, 89.37) \\
DeepSurv & 73.41 (71.61, 74.67) & \textbf{82.89 (82.40, 83.03)} & 68.38 (64.36, 70.95) & \textbf{90.11 (89.55, 90.43)} \\
DeepHit  & 74.25 (71.68, 74.71) & 82.84 (82.63, 83.01) & 66.84 (65.90, 68.28) & 89.82 (89.44, 90.13) \\
VSI      & 69.64 (54.03, 72.63) & 82.06 (78.92, 82.67) & 53.04 (45.10, 69.71) & 89.39 (88.27, 89.94) \\
SurVED   & 74.17 (70.46, 74.80) & 82.84 (81.07, 83.01) & 70.06 (67.05, 71.65) & 89.79 (88.95, 90.11) \\ 
\bottomrule
\end{tabular}}
\end{sc}
\end{footnotesize}
\end{center}
\vskip -0.1in
\end{table}
\begin{table}[t!]
\caption{The $\alpha$-Deviation values of the compared models on the four datasets. Numbers show the median, the 2.5\%, and the 97.5\% quantiles of 100-folds. All values are scaled by a factor of $10^{2}$. The lowest numerical value in each dataset is boldfaced.}
\label{tbl:comparison_with_other_models_CID_a_div}
\vskip 0.15in
\begin{center}
\begin{footnotesize}
\begin{sc}
\resizebox{\textwidth}{!}{%
\begin{tabular}{lcccc}
\toprule
& NWTCO & FLCHAIN & METABRIC & SUPPORT  \\
\midrule
CPH      & 1.81 (1.77, 1.85) & 4.65 (4.63, 4.68) & 4.75 (4.26, 5.17) & 1.26 (1.11, 1.35) \\
RSF      & 1.69 (1.64, 1.76) & 4.57 (4.52, 4.61) & 3.59 (3.20, 3.96) & 2.31 (2.06, 2.53) \\ 
DATE     & 1.47 (1.33, 1.61) & \textbf{4.30 (4.21, 4.45)} & 1.92 (1.07, 3.25) & 2.44 (2.00, 2.80) \\
DeepSurv & 1.72 (1.53, 1.84) & 4.67 (4.63, 4.72) & 3.89 (2.14, 4.70) & 1.18 (1.00, 1.35) \\
DeepHit  & 1.71 (1.62, 1.78) & 4.64 (4.59, 4.68) & 2.87 (2.20, 3.55) & \textbf{0.85 (0.73, 0.93)} \\
VSI      & \textbf{1.32 (0.52, 1.67)} & 4.62 (4.21, 4.77) & \textbf{1.73 (0.11, 4.14)} & 1.16 (0.76, 1.93) \\
SurVED   & 1.78 (1.53, 1.83) & 4.67 (4.47, 4.72) & 3.28 (2.31, 4.20) & 0.87 (0.61, 1.11) \\ 
\bottomrule
\end{tabular}}
\end{sc}
\end{footnotesize}
\end{center}
\vskip -0.1in
\end{table}

The Deep learning models outperformed classical models by a substantial margin on the SUPPORT dataset. To understand this notable difference and to explore how the models behave under different levels of censoring and dataset sizes, the following section employs the C-index decomposition to investigate the models' performances across various conditions simulated using the SUPPORT dataset.

\subsection{The Effect of Censoring and Size}
\label{ssct:Events Percentage Effect}
\begin{figure}[H]
     \centering
     \begin{subfigure}[b]{0.33\textwidth}
         \centering
         \includegraphics[width=\textwidth]{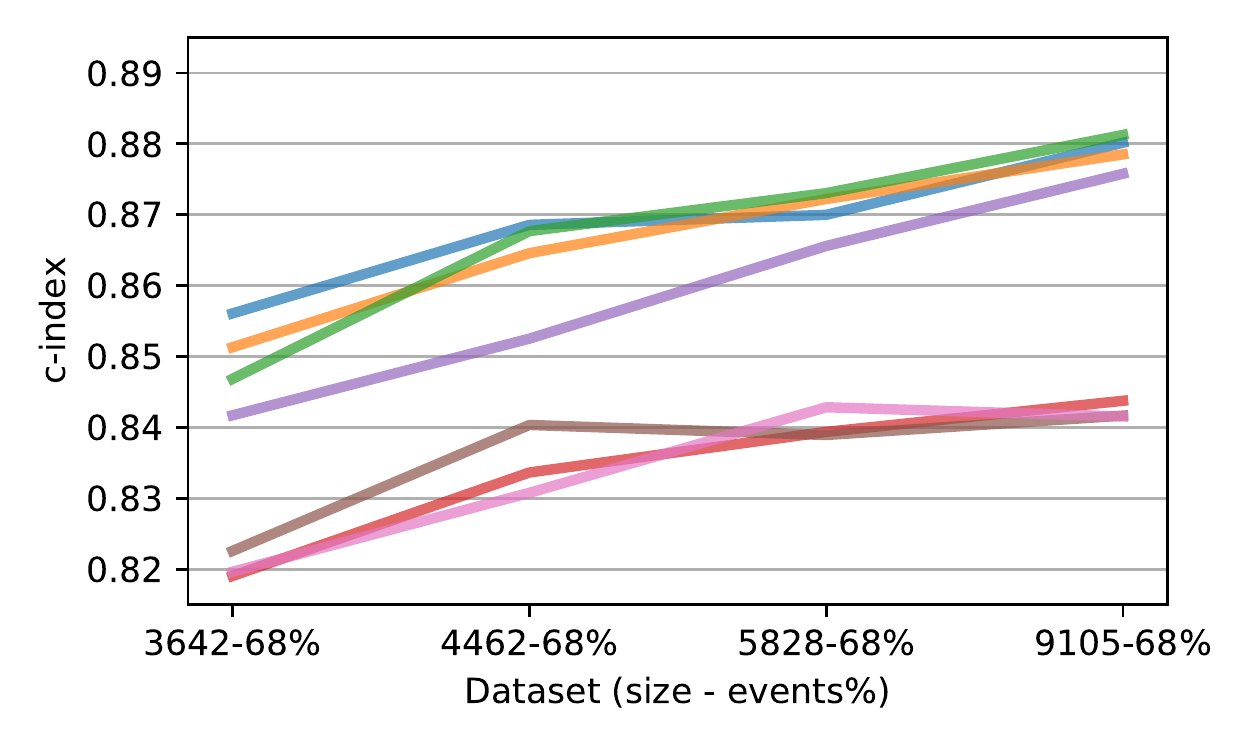}
         \caption{Size Only}
         \label{fig:c_size_only}
     \end{subfigure}%
     ~
     \begin{subfigure}[b]{0.33\textwidth}
         \centering
         \includegraphics[width=\textwidth]{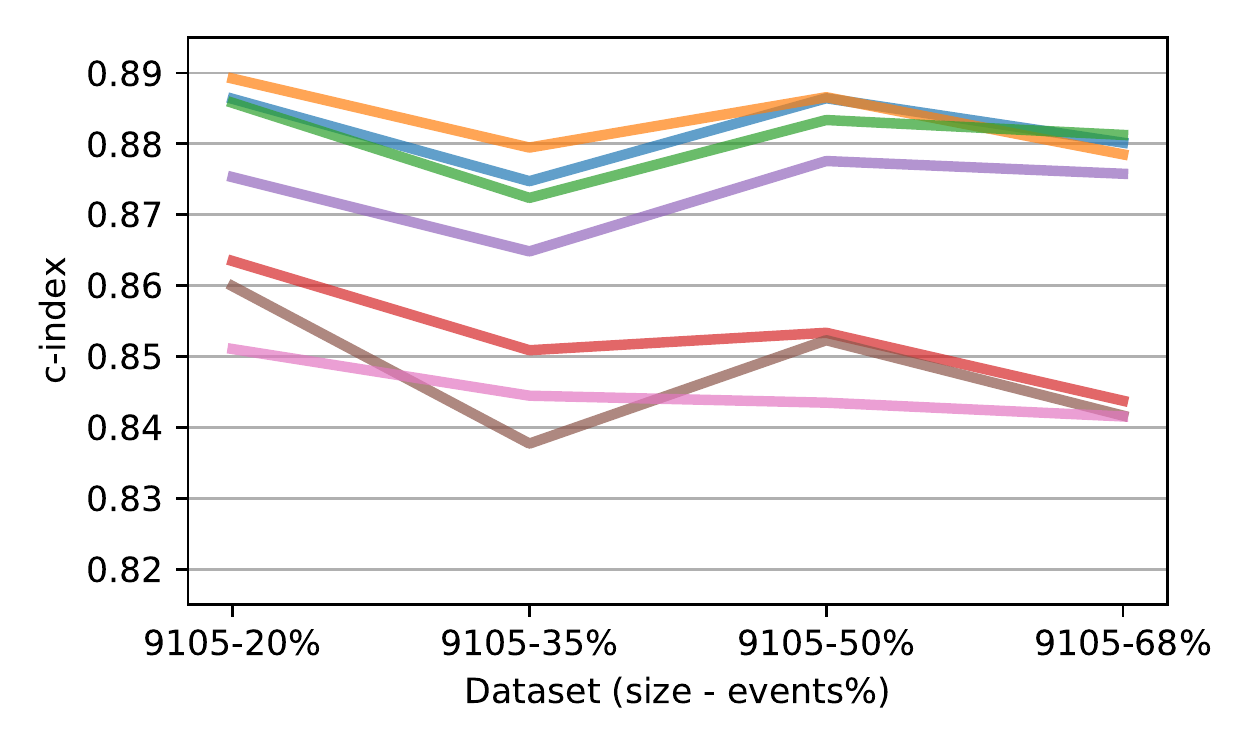}
         \caption{Censoring Only}
         \label{fig:c_censoring_only}
     \end{subfigure}%
     ~
     \begin{subfigure}[b]{0.33\textwidth}
         \centering
         \includegraphics[width=\textwidth]{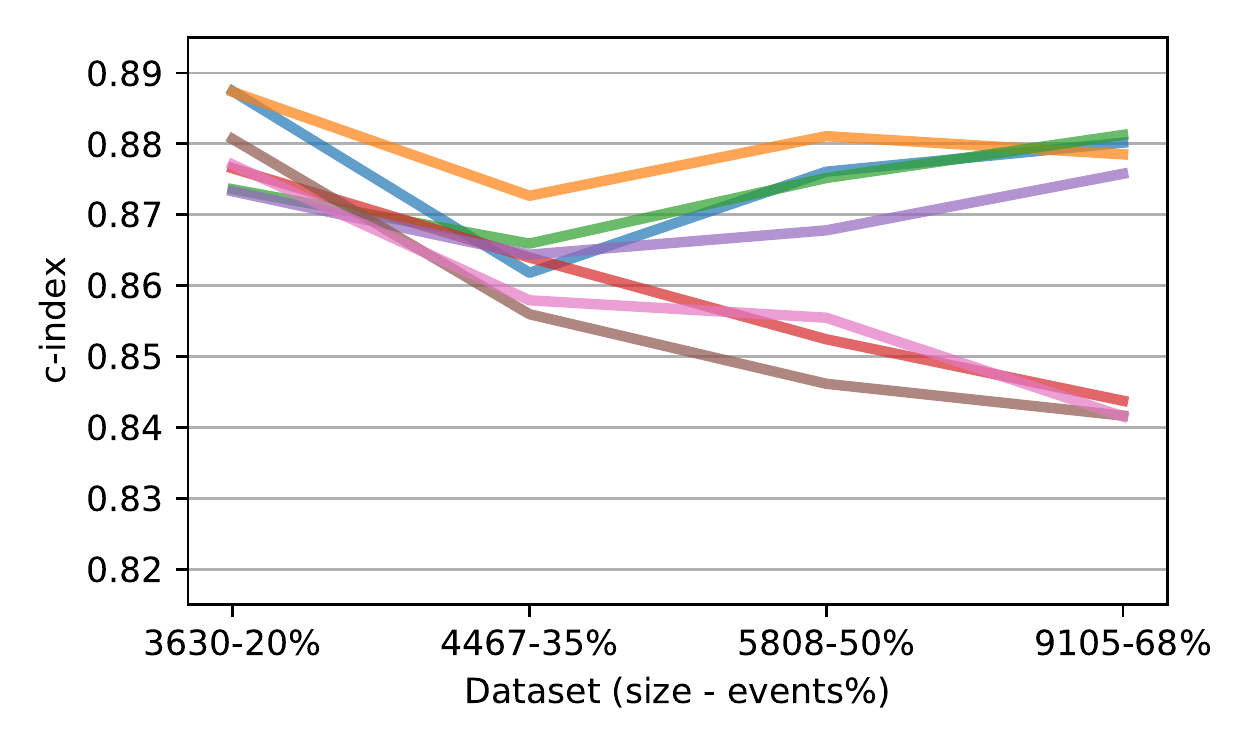}
         \caption{Size and Censoring}
         \label{fig:c_censoring_and_size}
     \end{subfigure}
     \hfill
     \begin{subfigure}[b]{0.8\textwidth}
         \centering
         \includegraphics[width=\textwidth]{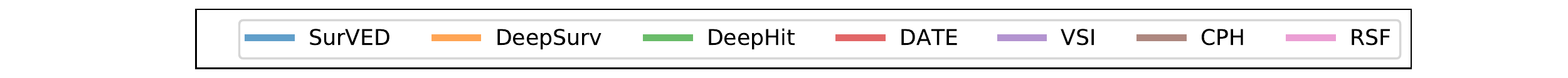}
     \end{subfigure}
        \caption{The change of $CI$ as the size of the dataset and the ratio of events change. The x-axis shows the sizes of the datasets and percentages of the events (for the SUPPORT dataset) in the three experiments.}
        \label{fig:support_uniform_c_all}
\end{figure}

Among the datasets utilized in this paper, the SUPPORT dataset is the largest and has the highest proportion of events. 
This characteristic allowed us to investigate the impact of varying the censoring and the dataset size across three different dimensions. 
Originally, the dataset contained 9,105 examples, with 6,201 observed events and 2,904 censored cases, resulting in $68\%$ events, and $32\%$ censored cases.
In the first experiment (Size Only), we varied the dataset size by randomly removing examples while keeping the censoring level fixed.
This resulted in four datasets with different sizes (3,642, 4,462, 5,828, and 9,105) and approximately the same event percentage of $68\%$.
In the second (Censoring Only), we varied the censoring level by randomly censoring observed events while maintaining the size. 
This resulted in four datasets of the same size (9,105) with varying event percentages ($20\%$, $35\%$, $50\%$, and $68\%$).
Lastly, in the third experiment (Size and Censoring), we simultaneously varied both dataset size and censoring level, by randomly dropping observed event examples.
This resulted in four datasets with different censoring levels (events percentages) ($20\%$, $35\%$, $50\%$, and $68\%$) and different sizes (3,630, 4,467, 5,808, and 9,105) respectively.
The models were trained and tested on each of the four datasets in each experiment, and Fig.~\ref{fig:support_uniform_c_all} illustrates how the C-indices for the models changed with varying dataset sizes and fractions of event cases (different levels of censoring). It is worth noting that the right-hand side of the three figures~\ref{fig:c_size_only}, \ref{fig:c_censoring_only}, and \ref{fig:c_censoring_and_size} is the performance of the models on the original SUPPORT dataset. 

Two distinct types of behaviors can be observed in these experiments (see Fig.~\ref{fig:support_uniform_c_all}): One related to the group \{SurVED, DeepSurv, DeepHit, VSI\}, i.e., the deep learning models except for DATE, and one related to the group \{DATE, CPH, RSF\}, i.e., the classical models plus DATE.
In the first experiment, Figure~\ref{fig:c_size_only}, where only the dataset size was changed, all the models improved in C-index performance as the dataset size increased. However, they maintained their relative differences between the two groups.
In the second experiment, Figure~\ref{fig:c_censoring_only}, where only the censoring level was varied, the models' performances remained relatively constant, with DATE and the classical models exhibiting a slight drop in the C-index performances.

The most intriguing result was obtained in the third experiment, Figure~\ref{fig:c_censoring_and_size}, where classical models behaved unexpectedly when both the size and the censoring level of the dataset were varied. 
The Deep learning models maintained a constant C-index performance as the data set size and the percentage of the observed events both decreased (reading Figure~\ref{fig:c_censoring_and_size} from right to left). In contrast, DATE and the classical models' performance improved eventually reaching a point where, in the extreme case of the smallest dataset and the lowest event percentage (the left-hand side of Figure~\ref{fig:c_censoring_and_size}), all models performed similarly.

To better understand these trends in the behavior concerning changes in censoring levels and dataset size, the performance of the models was further examined using the C-index Decomposition. The aim was to shed light on the underlying reasons behind such differences in behavior.

\begin{figure}[H]
     \centering
     \begin{subfigure}[b]{0.33\textwidth}
         \centering
         \includegraphics[width=\textwidth]{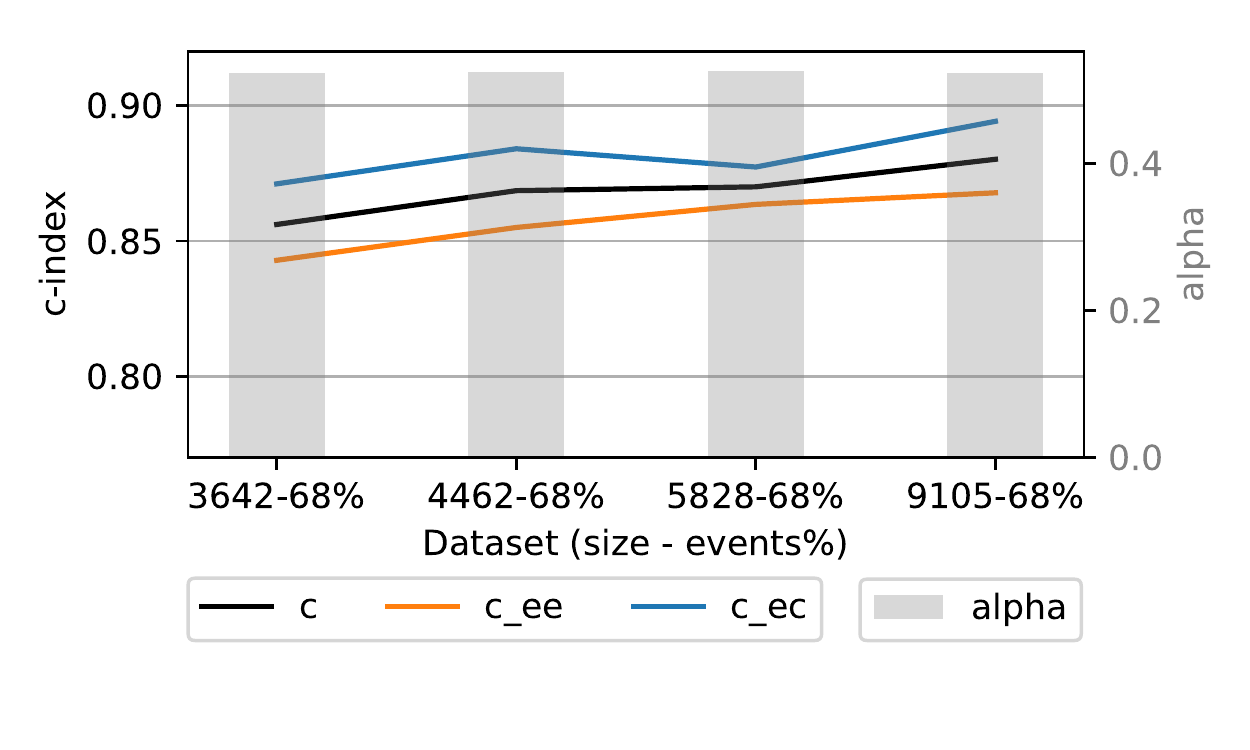}
         \vspace{-1.8\baselineskip}
         \caption{SurVED-Size Only}
         \label{fig:SurVED_size_only}
     \end{subfigure}%
     ~
     \begin{subfigure}[b]{0.33\textwidth}
         \centering
         \includegraphics[width=\textwidth]{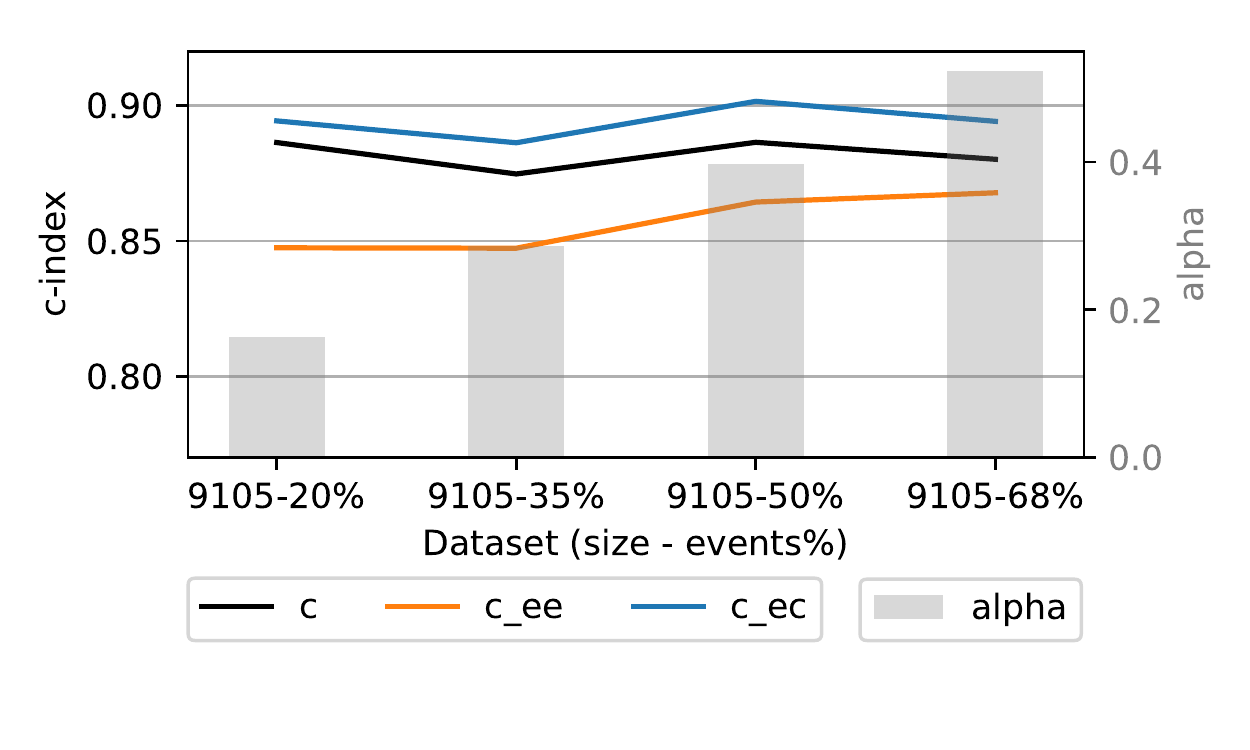}
         \vspace{-1.8\baselineskip}
         \caption{SurVED-Censoring Only}
         \label{fig:SurVED_censoring_only}
     \end{subfigure}%
     ~
     \begin{subfigure}[b]{0.33\textwidth}
         \centering
         \includegraphics[width=\textwidth]{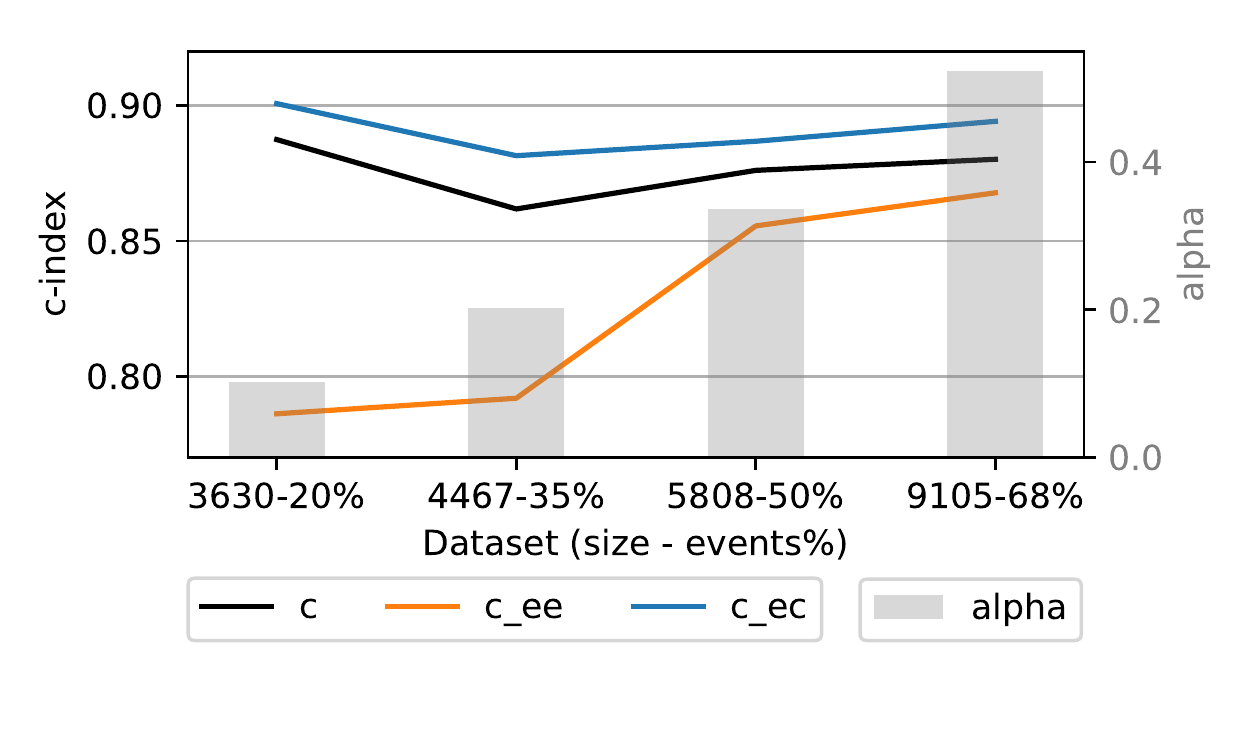}
         \vspace{-1.8\baselineskip}
         \caption{SurVED-Censoring and Size}
         \label{fig:SurVED_censoring_and_size}
     \end{subfigure}
     \hfill
    \begin{subfigure}[b]{0.33\textwidth}
         \centering
         \includegraphics[width=\textwidth]{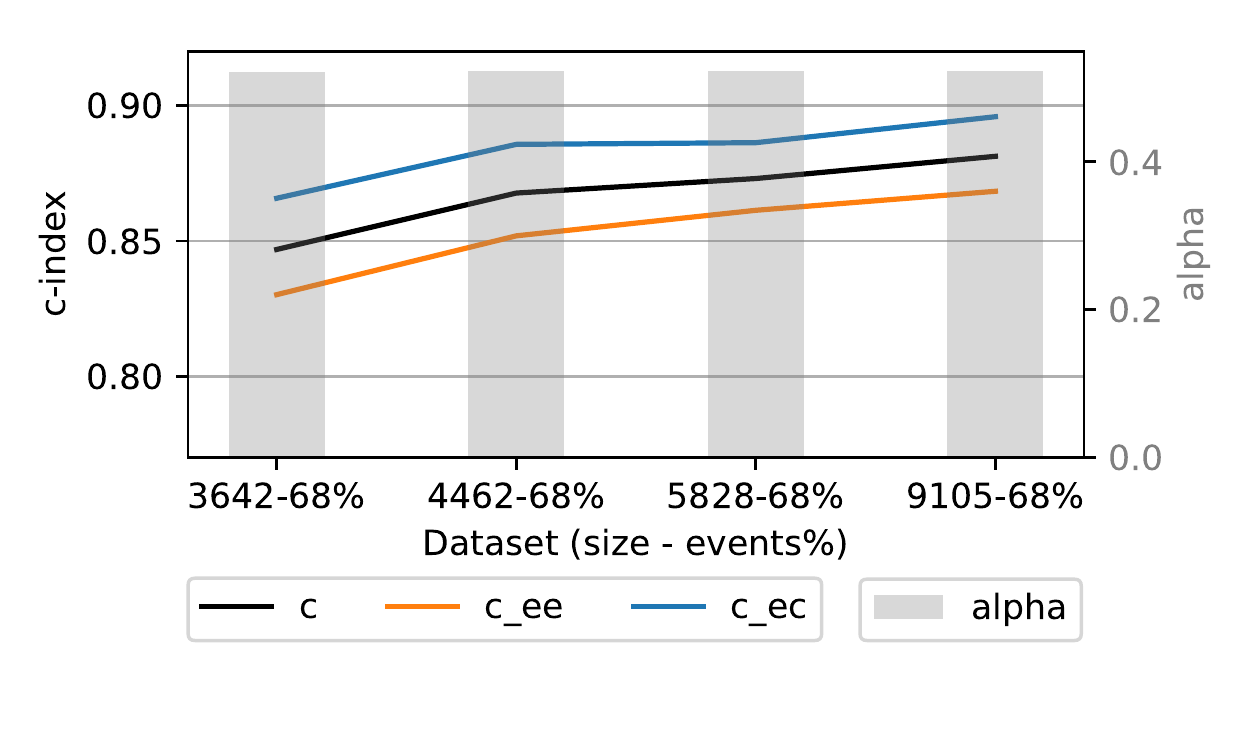}
         \vspace{-1.8\baselineskip}
         \caption{DeepHit-Size Only}
         \label{fig:DeepHit_size_only}
     \end{subfigure}%
     ~
     \begin{subfigure}[b]{0.33\textwidth}
         \centering
         \includegraphics[width=\textwidth]{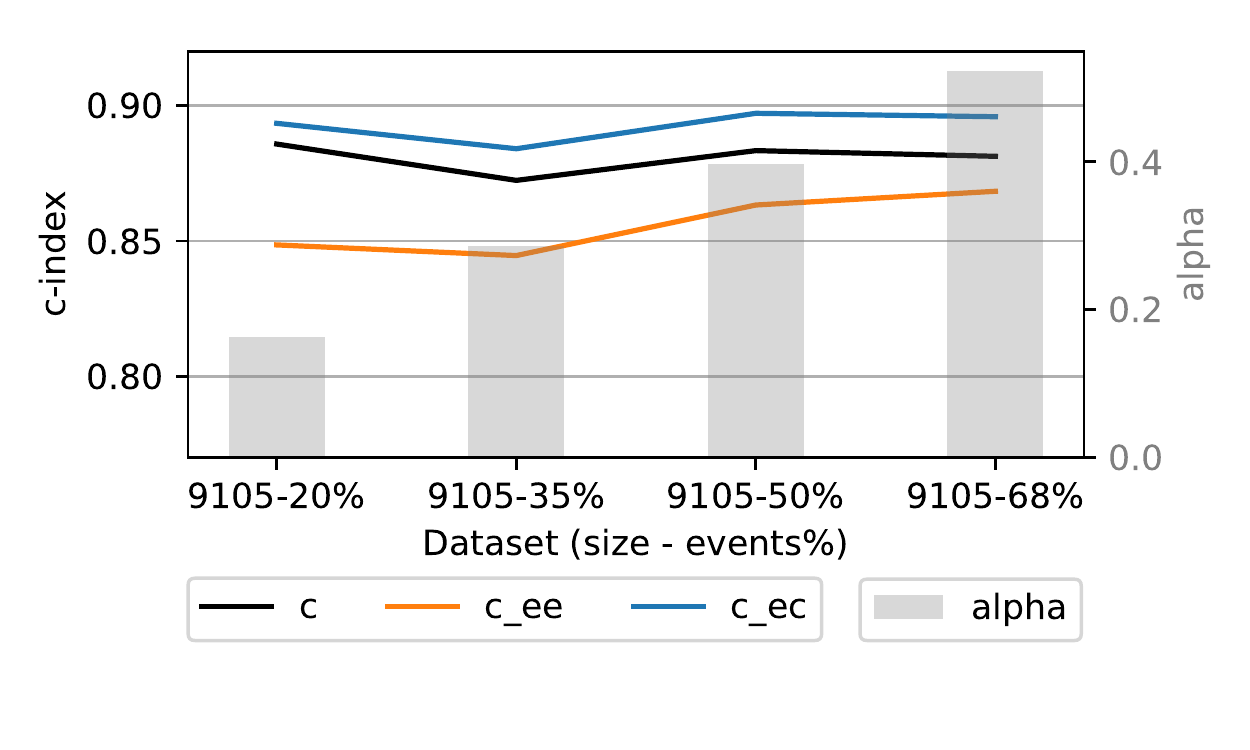}
         \vspace{-1.8\baselineskip}
         \caption{DeepHit-Censoring Only}
         \label{fig:DeepHit_censoring_only}
     \end{subfigure}%
     ~
     \begin{subfigure}[b]{0.33\textwidth}
         \centering
         \includegraphics[width=\textwidth]{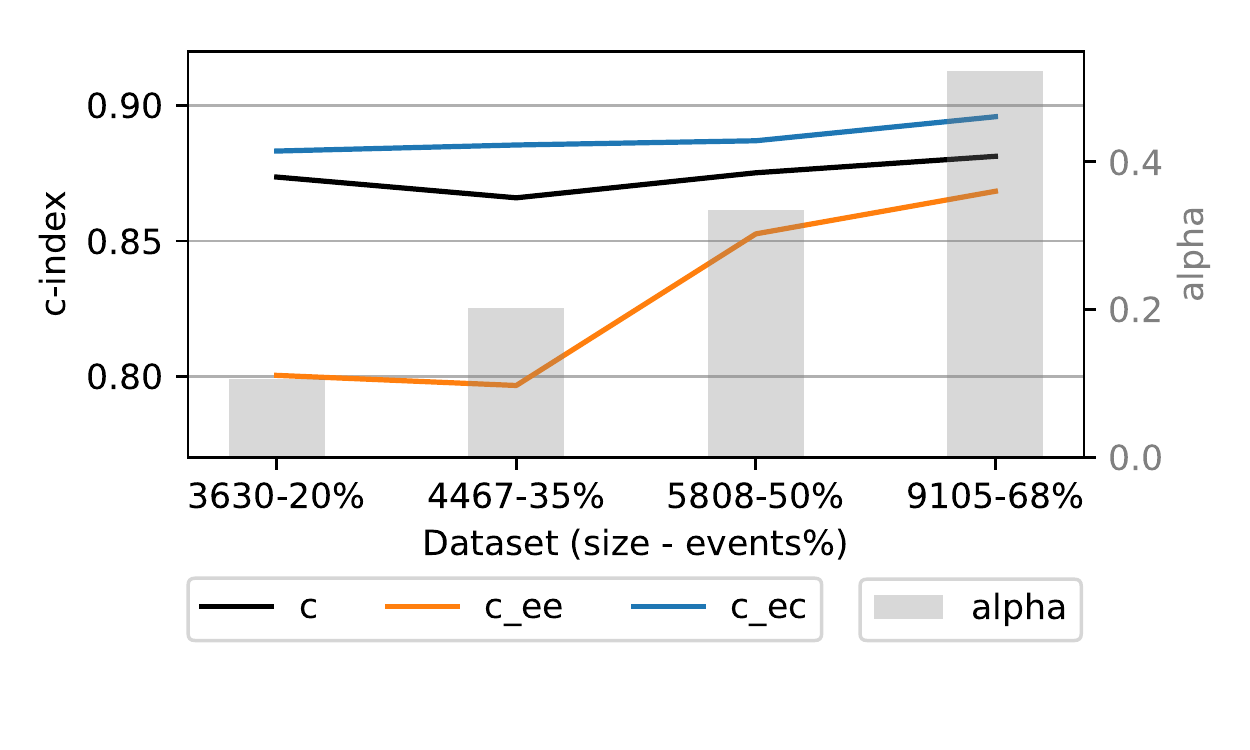}
         \vspace{-1.8\baselineskip}
         \caption{DeepHit-Censoring and Size}
         \label{fig:DeepHit_censoring_and_size}
     \end{subfigure}
     \begin{subfigure}[b]{0.33\textwidth}
         \centering
         \includegraphics[width=\textwidth]{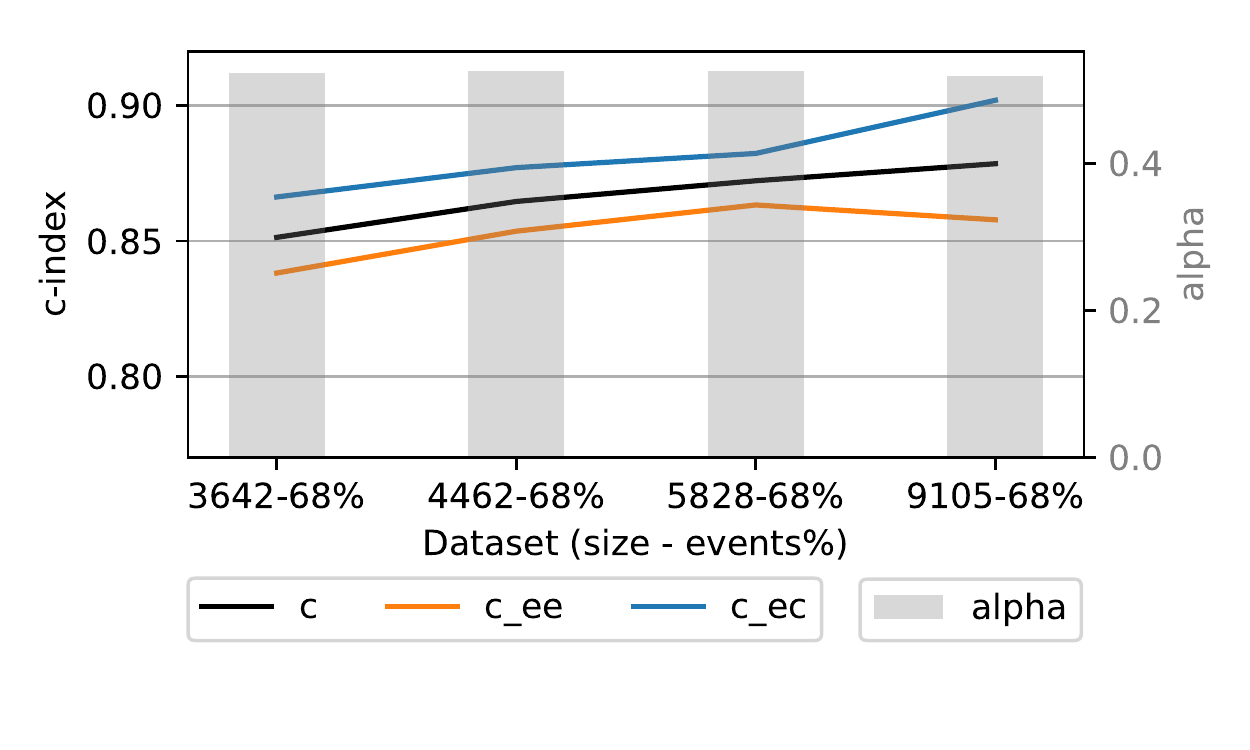}
         \vspace{-1.8\baselineskip}
         \caption{DeepSurv-Size Only}
         \label{fig:DeepSurv_size_only}
     \end{subfigure}%
     ~
     \begin{subfigure}[b]{0.33\textwidth}
         \centering
         \includegraphics[width=\textwidth]{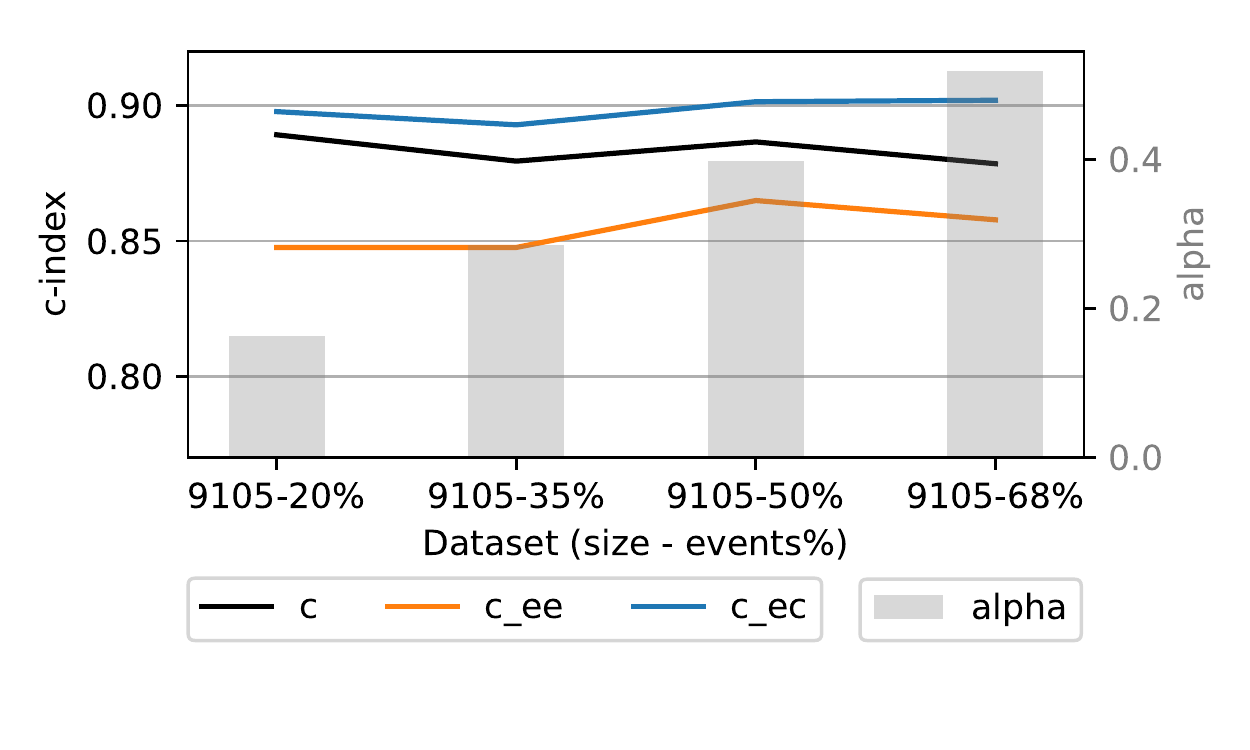}
         \vspace{-1.8\baselineskip}
         \caption{DeepSurv-Censoring Only}
         \label{fig:DeepSurv_censoring_only}
     \end{subfigure}%
     ~
     \begin{subfigure}[b]{0.33\textwidth}
         \centering
         \includegraphics[width=\textwidth]{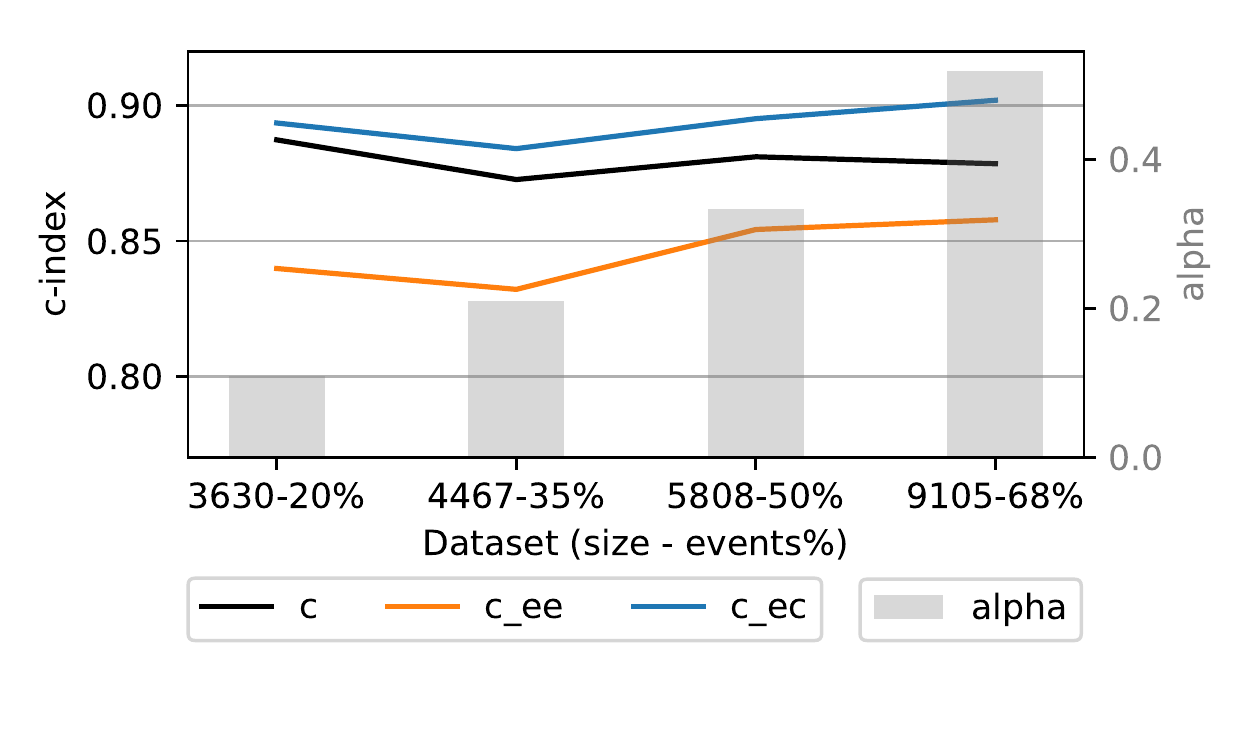}
         \vspace{-1.8\baselineskip}
         \caption{DeepSurv-Censoring and Size}
         \label{fig:DeepSurv_censoring_and_size}
     \end{subfigure}
     \begin{subfigure}[b]{0.33\textwidth}
         \centering
         \includegraphics[width=\textwidth]{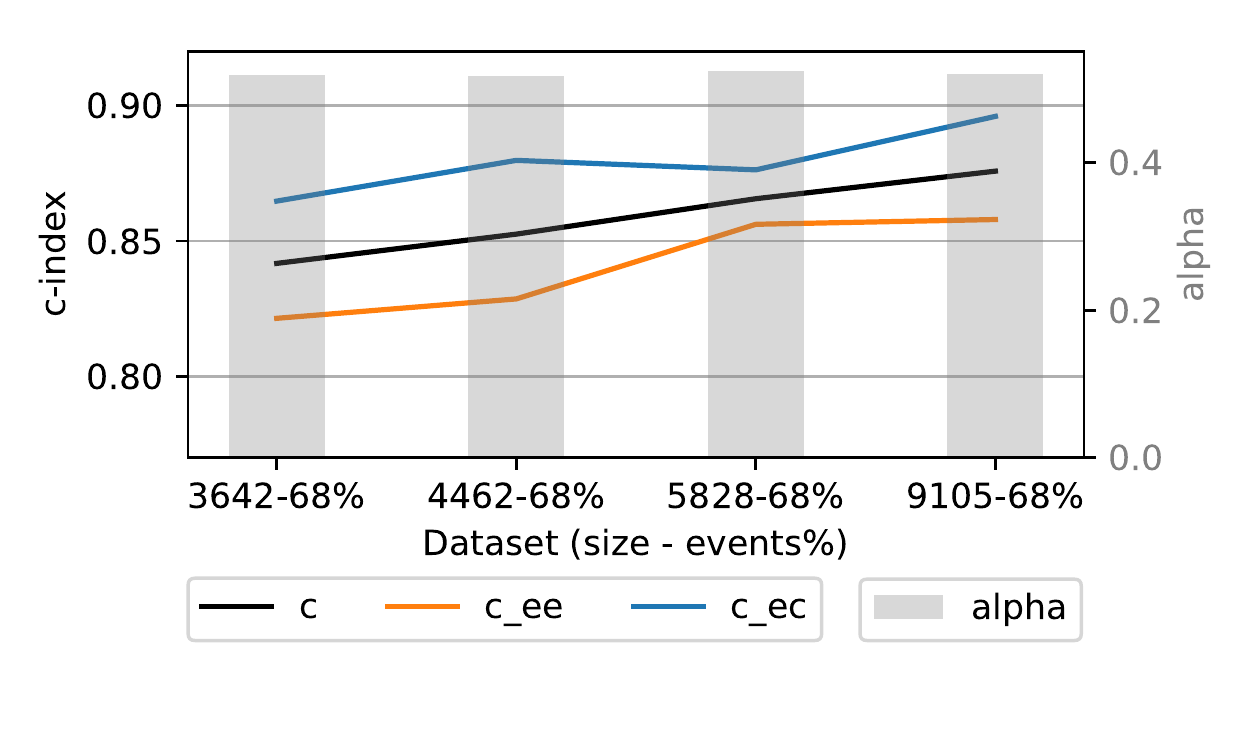}
         \vspace{-1.8\baselineskip}
         \caption{VSI-Size Only}
         \label{fig:VSI_size_only}
     \end{subfigure}%
     ~
     \begin{subfigure}[b]{0.33\textwidth}
         \centering
         \includegraphics[width=\textwidth]{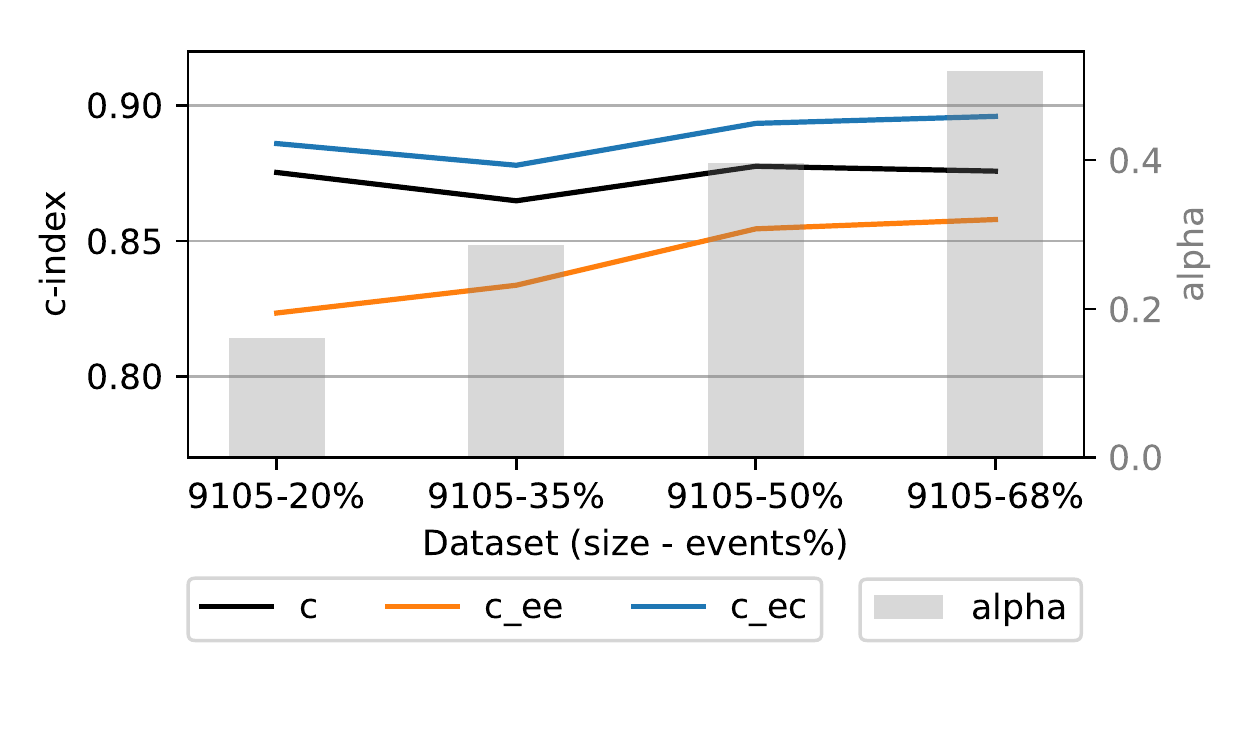}
         \vspace{-1.8\baselineskip}
         \caption{VSI-Censoring Only}
         \label{fig:VSI_censoring_only}
     \end{subfigure}%
     ~
     \begin{subfigure}[b]{0.33\textwidth}
         \centering
         \includegraphics[width=\textwidth]{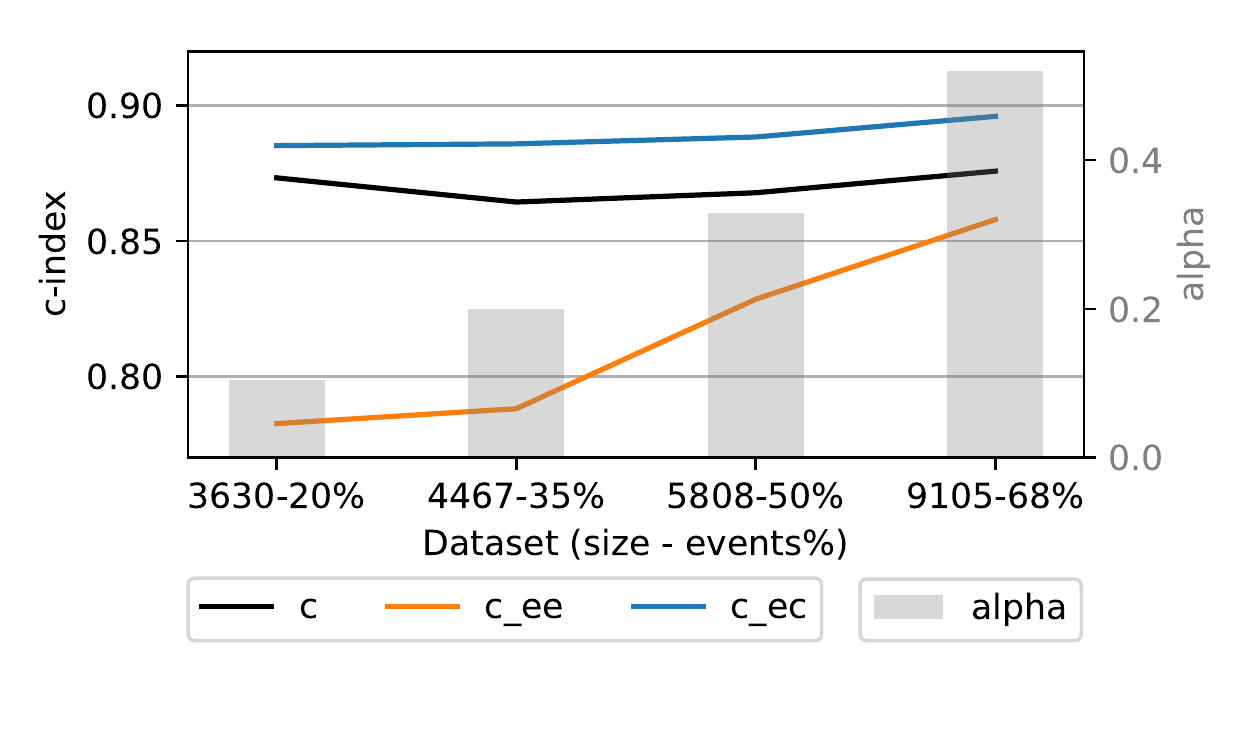}
         \vspace{-1.8\baselineskip}
         \caption{VSI-Censoring and Size}
         \label{fig:VSI_censoring_and_size}
     \end{subfigure}
     \begin{subfigure}[b]{0.33\textwidth}
         \centering
         \includegraphics[width=\textwidth]{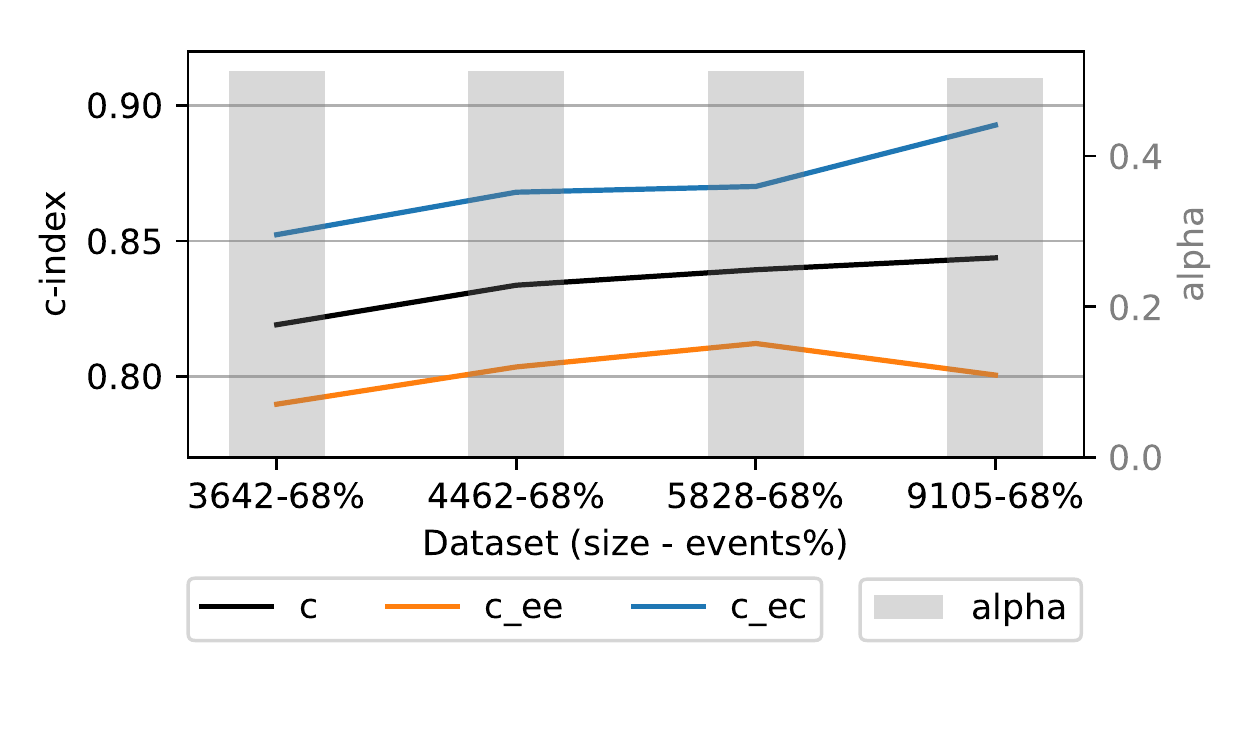}
         \vspace{-1.8\baselineskip}
         \caption{DATE-Size Only}
         \label{fig:DATE_size_only}
     \end{subfigure}%
     ~
     \begin{subfigure}[b]{0.33\textwidth}
         \centering
         \includegraphics[width=\textwidth]{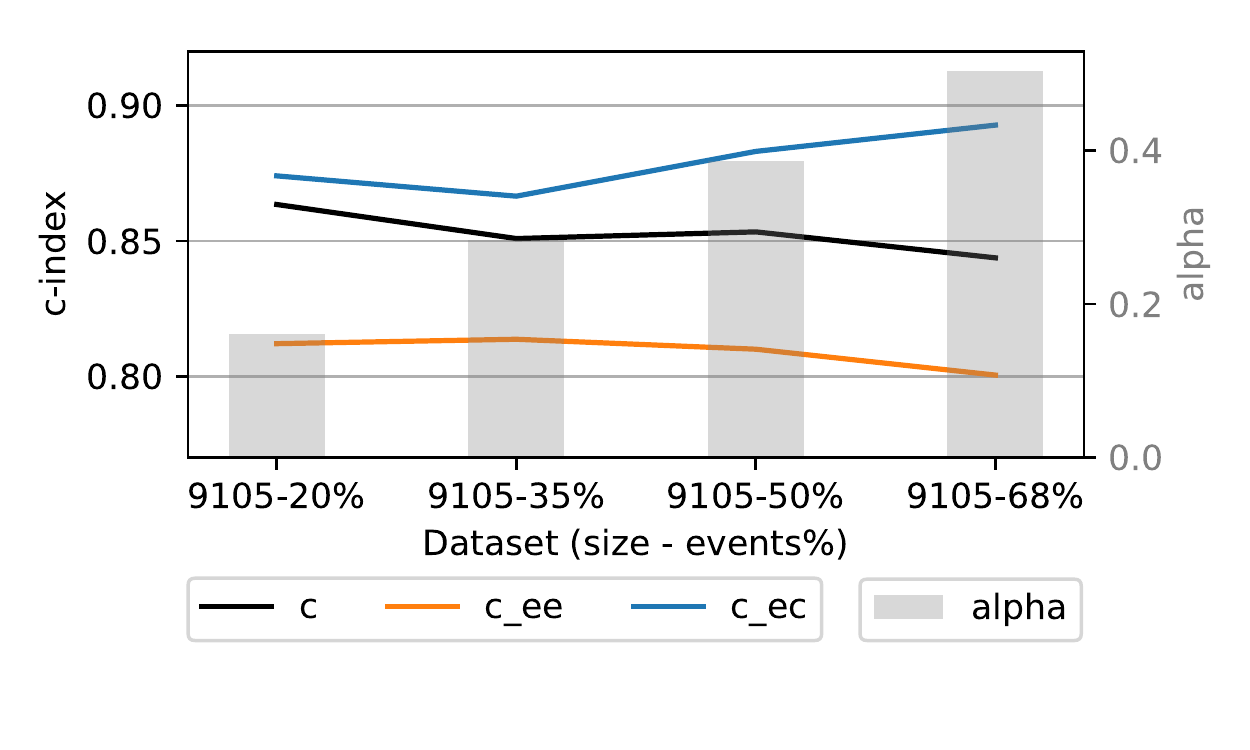}
         \vspace{-1.8\baselineskip}
         \caption{DATE-Censoring Only}
         \label{fig:DATE_censoring_only}
     \end{subfigure}%
     ~
     \begin{subfigure}[b]{0.33\textwidth}
         \centering
         \includegraphics[width=\textwidth]{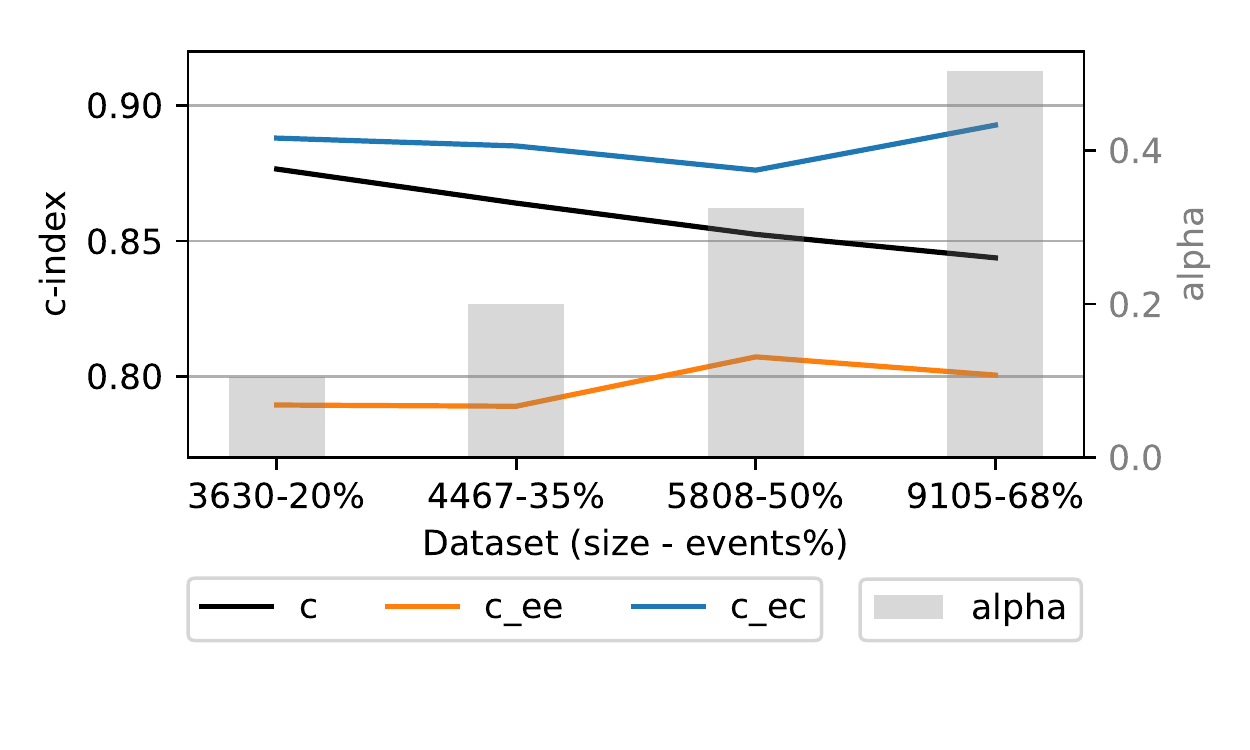}
         \vspace{-1.8\baselineskip}
         \caption{DATE-Censoring and Size}
         \label{fig:DATE_censoring_and_size}
     \end{subfigure}
     \begin{subfigure}[b]{0.33\textwidth}
         \centering
         \includegraphics[width=\textwidth]{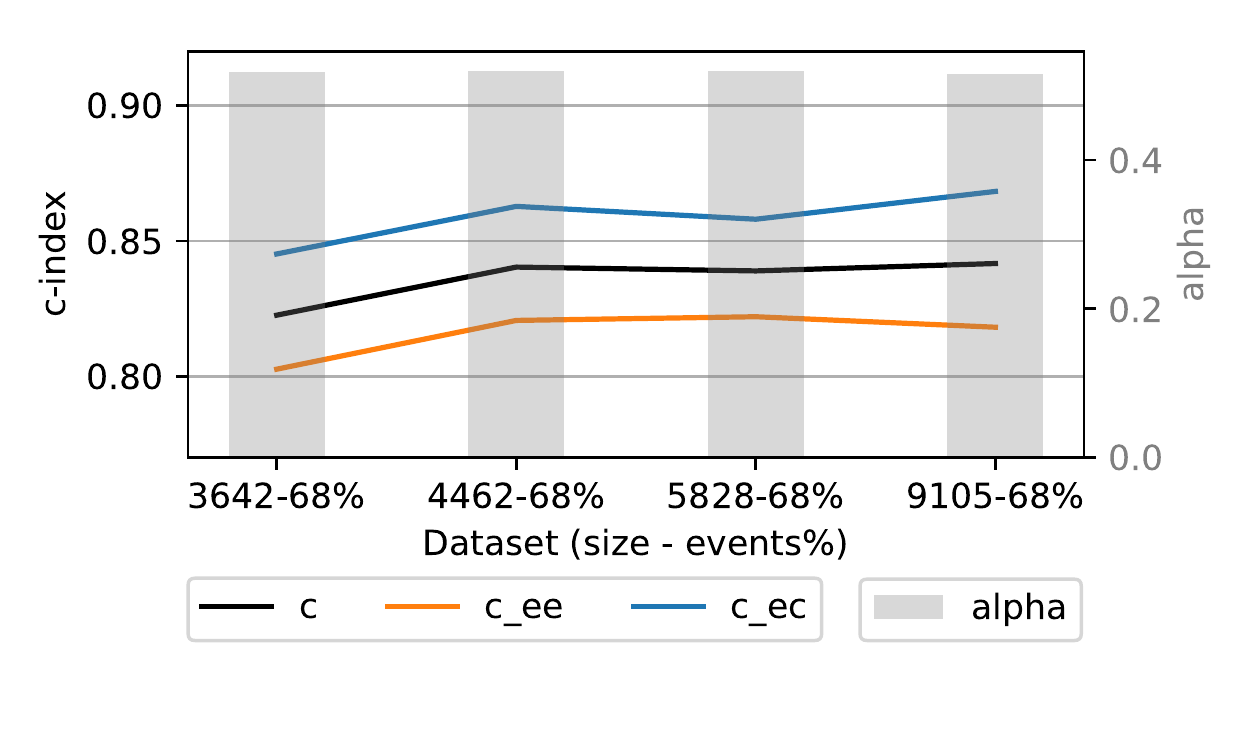}
         \vspace{-1.8\baselineskip}
         \caption{CPH-Size Only}
         \label{fig:CPH_size_only}
     \end{subfigure}%
     ~
     \begin{subfigure}[b]{0.33\textwidth}
         \centering
         \includegraphics[width=\textwidth]{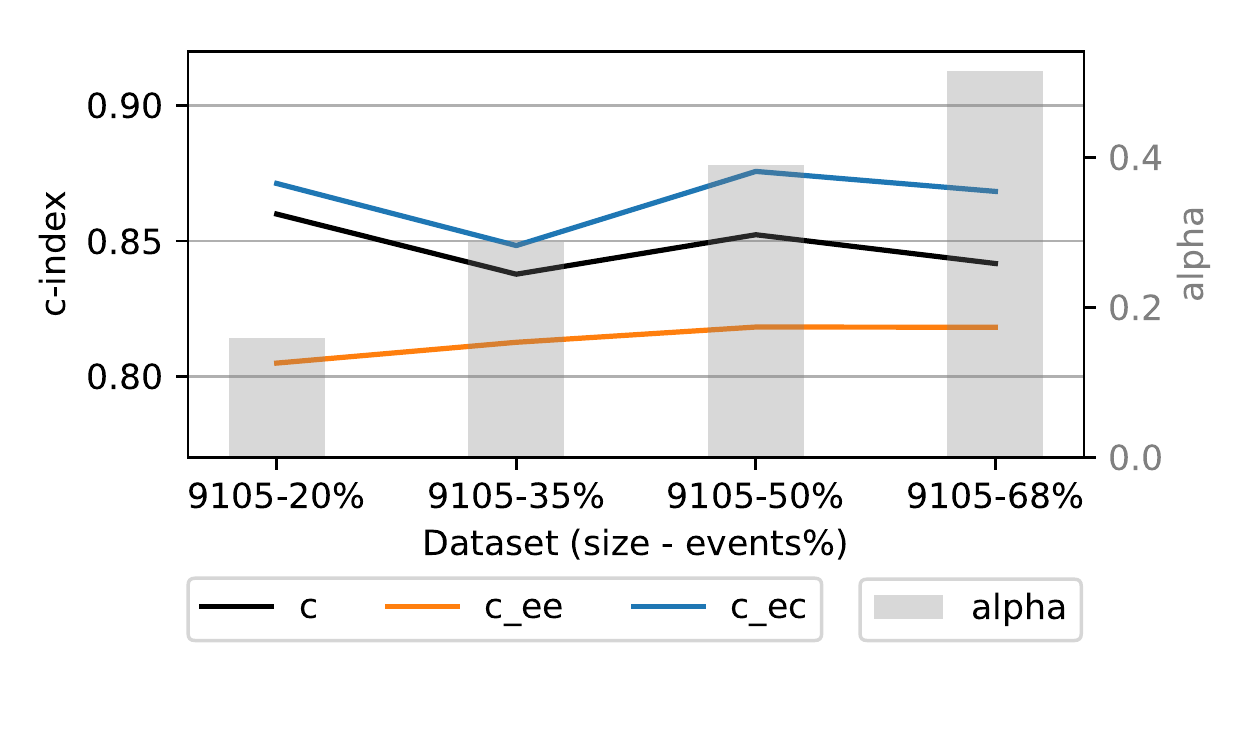}
         \vspace{-1.8\baselineskip}
         \caption{CPH-Censoring Only}
         \label{fig:CPH_censoring_only}
     \end{subfigure}%
     ~
     \begin{subfigure}[b]{0.33\textwidth}
         \centering
         \includegraphics[width=\textwidth]{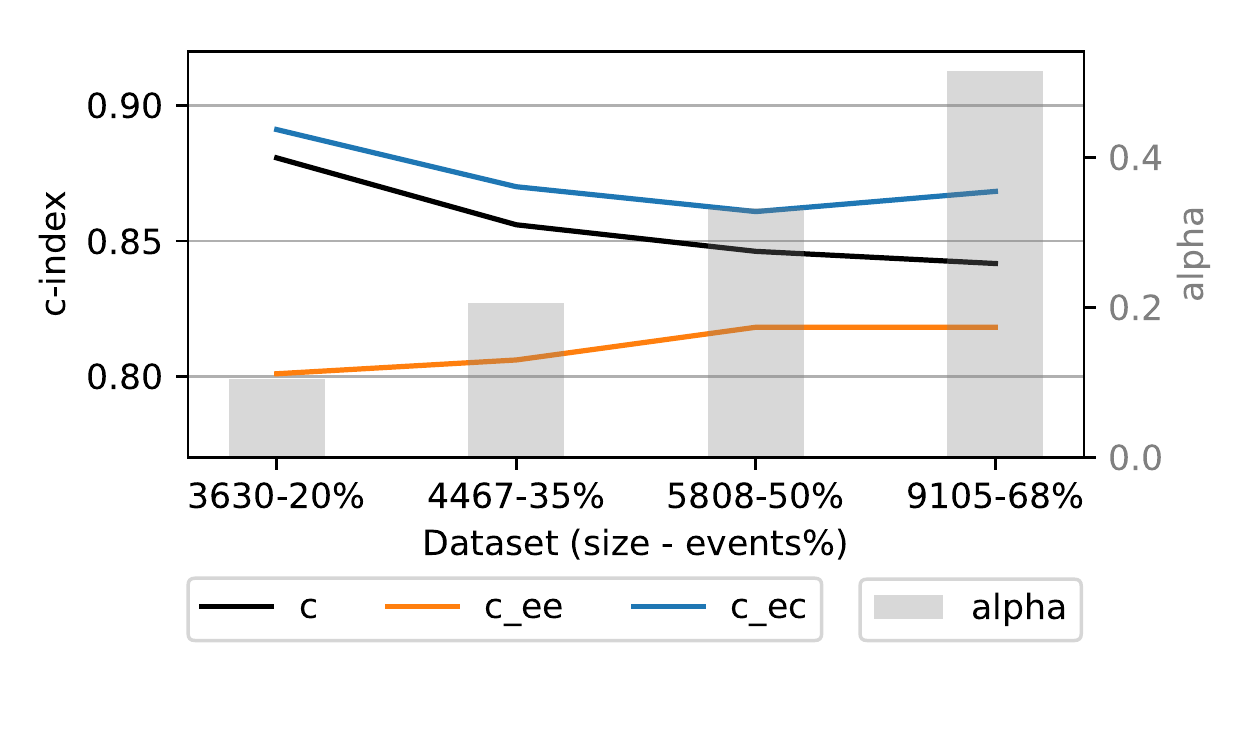}
         \vspace{-1.8\baselineskip}
         \caption{CPH-Censoring and Size}
         \label{fig:CPH_censoring_and_size}
     \end{subfigure}
     \begin{subfigure}[b]{0.33\textwidth}
         \centering
         \includegraphics[width=\textwidth]{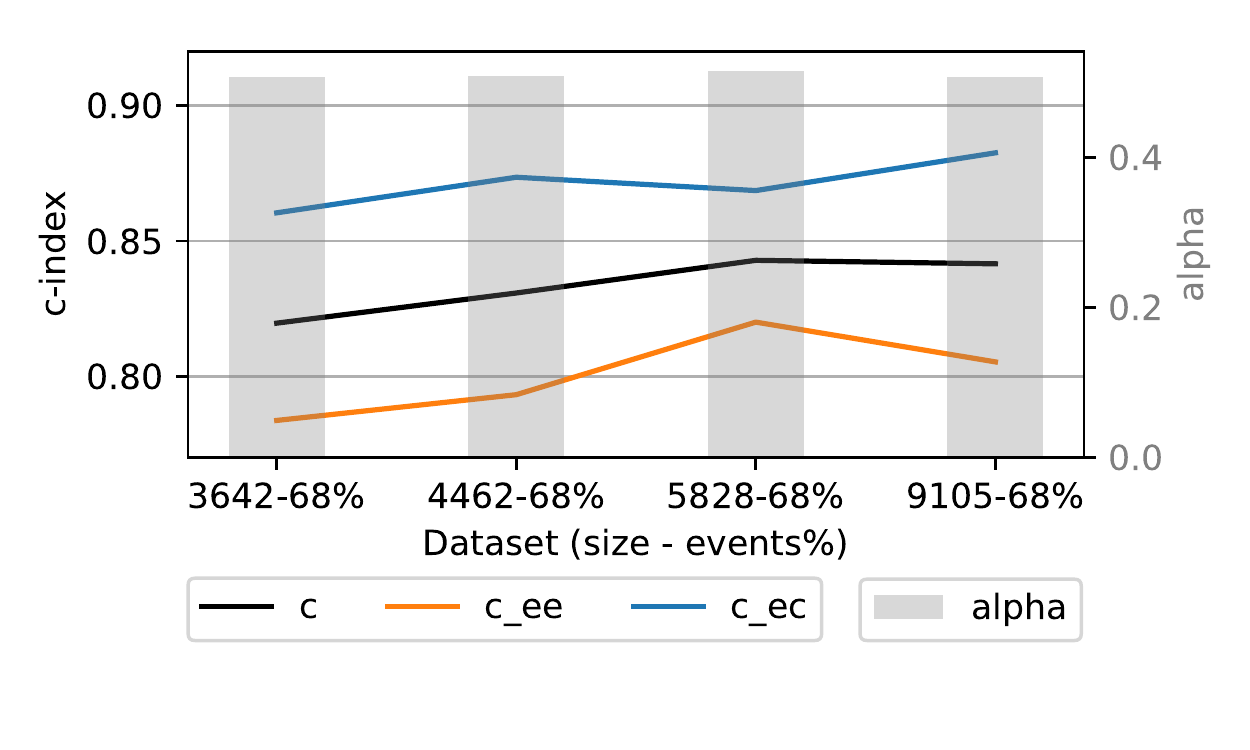}
         \vspace{-1.8\baselineskip}
         \caption{RSF-Size Only}
         \label{fig:RSF_size_only}
     \end{subfigure}%
     ~
     \begin{subfigure}[b]{0.33\textwidth}
         \centering
         \includegraphics[width=\textwidth]{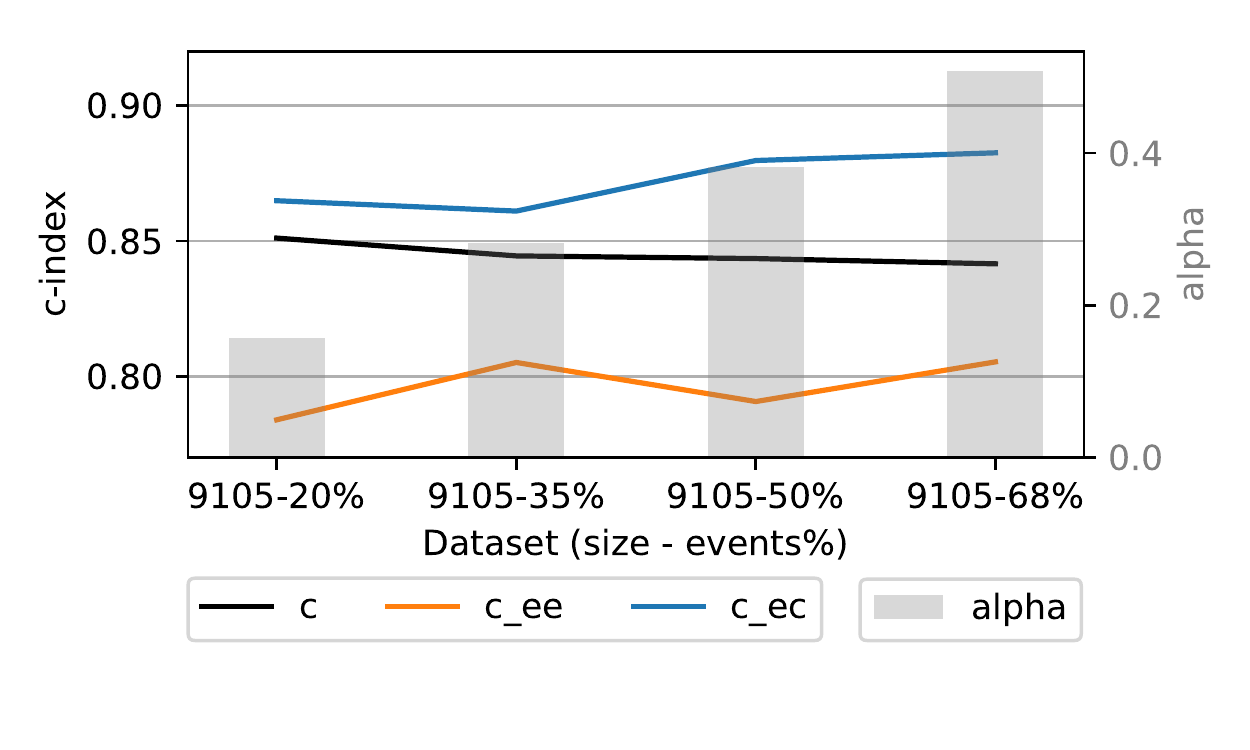}
         \vspace{-1.8\baselineskip}
         \caption{RSF-Censoring Only}
         \label{fig:RSF_censoring_only}
     \end{subfigure}%
     ~
     \begin{subfigure}[b]{0.33\textwidth}
         \centering
         \includegraphics[width=\textwidth]{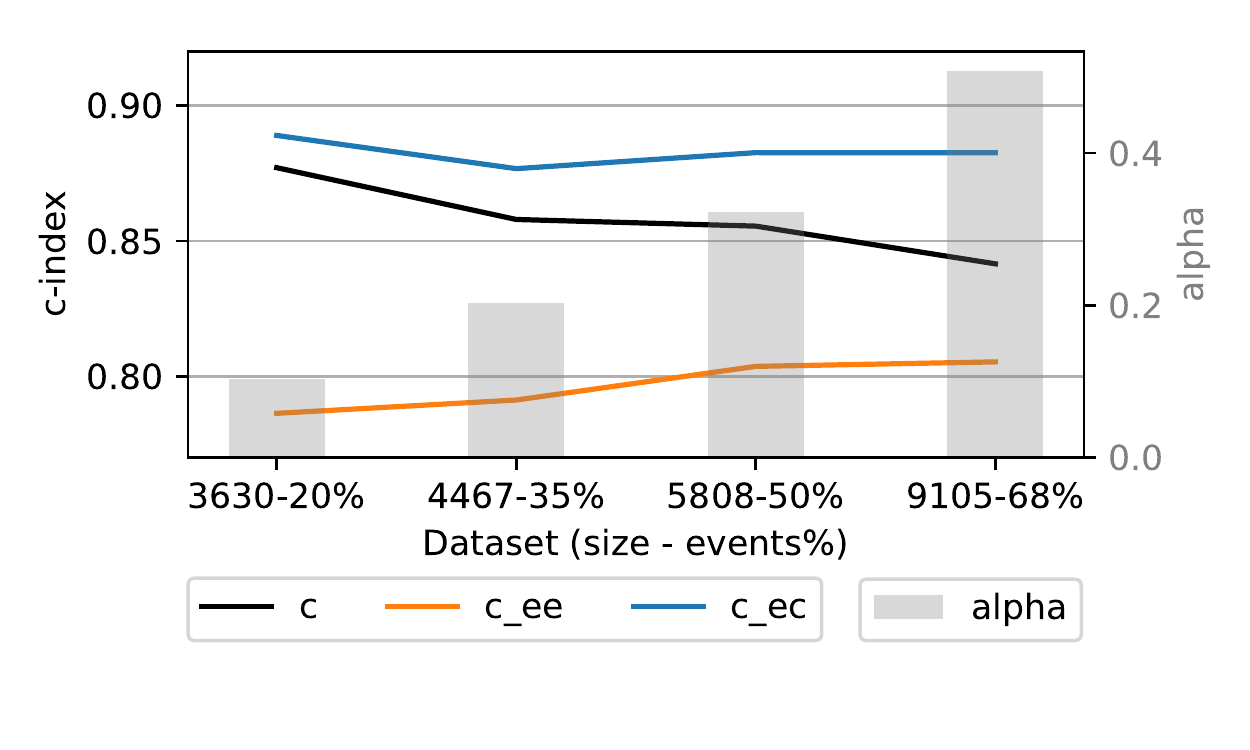}
         \vspace{-1.8\baselineskip}
         \caption{RSF-Censoring and Size}
         \label{fig:RSF_censoring_and_size}
     \end{subfigure}
        \caption{The change of $CI$, $CI_{ee}$, $CI_{ec}$, and $\alpha$ in eq.~(\ref{eq:CI_ee_ec}) as the ratio of events changes. The x-axis shows different percentages of events (for the SUPPORT dataset).}
        \label{fig:support_uniform_c_decomp_all}
\end{figure}

Figure~\ref{fig:support_uniform_c_decomp_all} shows the C-index decomposition of the seven models on SUPPORT datasets in the three experiments (varying the dataset size only, varying the censoring level only, and varying both the size and the censoring level). Two distinct trends in behavior are observed: one corresponding to classical models, CPH and RSF. The other one corresponds to the deep learning models except for DATE, which followed the classical models' behavior. Hence DATE will be included with the classical models when referring to the classical models' behavior below.

In the first experiment (the leftmost column in Figure~\ref{fig:support_uniform_c_decomp_all}), increasing the size of the dataset led to an increase in both the $CI_{ee}$ and $CI_{ec}$. Furthermore, keeping the percentage of the events fixed maintained similar values for the $\alpha$ term in the decomposition through the four datasets (approximately $0.5$). This balance in the $\alpha$ gave equal weight to the two terms in the C-index decomposition resulting in improvement in the total C-index for all models with increased dataset size.

In the second experiment (the middle column in Figure~\ref{fig:support_uniform_c_decomp_all}), keeping the size fixed and decreasing the censoring level (increasing events \%) slightly increased the $CI_{ee}$ performance for deep learning models and, to a lesser extent, for classical models. On the other hand, the $CI_{ec}$ stayed almost constant for deep learning models, with a slight increase for classical models. Nevertheless, changing the censoring level affected $\alpha$ changing the weighting on the two decomposition terms across four datasets. As a result, with smaller $\alpha$, the total C-index was mainly influenced by the $CI_{ec}$ at the high censoring level (low events \% to the left side of the figure), whereas $\alpha$ increases (hence the weight on the $CI_{ee}$) as the events percentage increase. This caused the total C-index to stay constant for deep learning models but slightly decreased for classical models.

In the third experiment, when changing the dataset's size and the censoring level (the column to the right in Figure~\ref{fig:support_uniform_c_decomp_all}), the impact became more pronounced. All the methods essentially achieved high C-indices at a high censoring level (low \% of events) and smaller dataset, resulting in very similar performances with respect to $CI$, $CI_{ee}$, and $CI_{ec}$. However, at such a high censoring level, the $\alpha$ term of the C-index is relatively small, which makes the C-index primarily influenced by the $CI_{ec}$ term with minimal contribution from the $CI_{ee}$ term. As the size increases and censoring decreases, the $\alpha$ value increases, giving more weight to the $CI_{ee}$ term. In this case, as the classical models did not exhibit improvements on the $CI_{ee}$, which remained almost the same as more events were added to the dataset, this caused the total C-index to decrease with the increasing weight on this term. In contrast, the deep learning models exhibited an increase in $CI_{ee}$, which kept the total C-index the same for all levels of censoring.

The main difference between the second and the third experiments lies in their approach to handling censoring. In the second experiment (Censoring Only), a fraction of the observed event examples are censored, while in the third experiment (Censoring and size) those observed event examples are entirely removed from the dataset. To achieve the same censoring percentage in the two scenarios, more event cases need to be removed in the third experiment compared to the ones that need to be censored in the second experiment. This results in that, for example, a dataset with $20\%$ events in the second experiment has 1,821 event cases compared to 726 event cases in a dataset with a similar event percentage in the third experiment. This explains the larger drop in performance in the $CI_{ee}$ in the third experiment which has less number of observed event cases.

\section{Conclusion}
\label{sct:conclusion}
In this work, we derived a decomposition of the C-index, separating it into two terms: one for ranking observed events, and another for ranking observed events versus censored cases. These terms are weighted by the parameter $\alpha$. The $\alpha$ factor expresses the contribution of the two parts for the total C-index and can be interpreted as a conditional probability for event-event pairs given that it is correctly ordered $P((ee)\ \mbox{pair} | \mbox{ordered pair})$. A model that perfectly orders the two types of pairs will have an optimal $\alpha$ factor ($\alpha^{*}$). Unbalanced models, i.e., models that are not equally good at ranking event-event pairs and event-censored pairs will deviate from this value. Based on this deviation from the $\alpha^{*}$, the $\alpha$-Deviation measure can assess how balanced a model is with respect to the ranking of the two groups of pairs.

SurVED is also proposed, a new approach for estimating the time-to-event distribution using a variational encoder-decoder with a Gaussian latent layer. In benchmark tests, SurVED performs significantly better than the two closely related methods, DATE and VSI, and achieves a comparable overall performance to DeepSurv and DeepHit.

Using the C-index decomposition, it was shown that in cases where models perform differently in terms of the $CI_{ee}$ and $CI_{ec}$, such differences often go unnoticed when evaluating the total C-index due to the averaging. Furthermore, it was demonstrated, using the SUPPORT dataset with varying censoring levels and dataset size, that all methods benefitted from increasing the dataset size. It was also shown that all methods have comparable performance in terms of the total C-index at a high censoring percentage and smaller dataset size, but all methods do better at ranking event-censored pairs compared to ranking event-event pairs. However, as the number of events grows, SurVED and the other deep learning models VSI, DeepSurv, and DeepHit are better than the other algorithms at improving their performance in ranking event-event pairs. This helped deep learning models maintain a constant C-index performance across different censoring levels in contrast to the classical models which suffered from a drop in the C-index. This explains the large magnitude of the difference between deep learning models and the classical models on the SUPPORT dataset.

This work focuses on analyzing the ranking performance of survival models using the C-index decomposition trying to get a better understanding of the strengths and weaknesses of models with respect to the different types of events and censored observations. Such understanding drawn from decomposition can help to design better objective functions of survival models which we leave for future work. Moreover, studying the relation between the decomposition terms and other evaluation metrics can potentially give more insights that help develop better survival models which we also leave for future work.  

\section{Declarations of interest}
The authors declare that they have no known competing financial interests or personal relationships that could have influenced the work reported in this paper.

\section{Acknowledgement}
This research was performed under the CAISR+ project funded by the Swedish Knowledge Foundation.

\bibliography{mybibfile}

\end{document}